\documentclass[11pt]{article}

\usepackage[final]{acl}

\usepackage{times}
\usepackage{latexsym}

\usepackage[T1]{fontenc}
\usepackage[utf8]{inputenc}

\usepackage{microtype}

\usepackage{inconsolata}

\usepackage{graphicx}

\usepackage{amsmath,amsfonts}
\usepackage{array}
\usepackage{booktabs} 
\usepackage{dsfont}
\usepackage{float}
\usepackage{makecell}
\usepackage{multirow}
\title{How Do Answer Tokens Read Reasoning Traces?\\ Self-Reading Patterns in Thinking LLMs for Quantitative Reasoning}

\author{
    Haoyang Chen$^\dagger$ \quad 
    Yi Liu$^\dagger$ \quad 
    Jianzhi Shao$^\ddagger$ \quad
    Tao Zhang$^\S$ \quad
    Chengfu Huo$^\ddagger$ \quad
    Wei Hu$^{\dagger,}$\thanks{\,\, Corresponding author} \\
    $^\dagger$ State Key Laboratory for Novel Software Technology, Nanjing University, China \\
    $^\ddagger$ Alibaba Group, China \qquad $^\S$ Ant Group, China \\
    \texttt{chen7haoyang@gmail.com, yiliu07.nju@gmail.com, sjz250796@taobao.com} \\
    \texttt{guyan.zt@antgroup.com, chengfu.huocf@taobao.com, whu@nju.edu.cn}
}

\begin{document}
\maketitle

\begin{abstract}
Thinking LLMs produce reasoning traces before answering. 
Prior activation steering work mainly targets on shaping these traces. 
It remains less understood how answer tokens actually read and integrate the reasoning to produce reliable outcomes. 
Focusing on quantitative reasoning, we analyze the answer-to-reasoning attention and observe a benign self-reading pattern aligned with correctness, characterized by a forward drift of the reading focus along the reasoning trace and a persistent concentration on key semantic anchors, whereas incorrect solutions exhibit diffuse and irregular attention pattern.
We interpret this as internal certainty during answer decoding, where the model commits to a viable solution branch and integrates key evidence. 
Following this, we propose a training-free steering method driven by Self-Reading Quality (SRQ) scores combining geometric metrics for process control with semantic metrics for content monitoring.
SRQ selects data to build steering vectors that guide inference toward benign self-reading and away from uncertain and disorganized reading. 
Experiments show that our method yields consistent accuracy gains.
\end{abstract}

%====================%
\section{Introduction}
\label{sec:intro}

Thinking LLMs (a.k.a.~reasoning models), such as DeepSeek-R1, GPT-5, and Gemini 3 series, exhibit strong quantitative reasoning abilities and typically generate a reasoning trace (e.g., separated by \texttt{</think>}) before the final answer \citep{thinking}.
To control these models, activation steering emerges as a powerful, training-free intervention method for modulating latent representations. 
While proven robust for general alignment tasks like eliciting honesty \citep{honest} and ensuring instruction following \citep{instruct}, recent work adapts steering techniques to optimize the reasoning traces of thinking LLMs for interpreting internal mechanisms \citep{understanding}, compressing verbose outputs \citep{ASE}, eliciting extended capabilities \citep{long_cot}, and calibrating reliability \citep{seal}.

Despite these efforts, \emph{it remains unclear how answer tokens actually read the reasoning trace}.
An inspiring analysis \citep{from} confirms the existence of attention links between answer tokens and reasoning tokens, but there is still a lack of exploration of how answer tokens navigates noise and leverages key information from the reasoning trace during decoding.
This issue is particularly important in thinking LLMs where reasoning chains often span thousands of tokens. 

To understand how answer tokens read and utilize reasoning tokens, we investigate the self-reading behavior of thinking LLMs by conducting an attention analysis on quantitative reasoning tasks.
We identify a stable, structured reading pattern emerging in the middle-to-late layers, correlating strongly with correctness.
In correct samples, rather than scanning the entire trace indiscriminately, the answer tokens exhibit a \emph{benign} self-reading mode characterized by two features: (i) a progressive forward shift of the attention centroid (or reading focus) toward later reasoning positions as decoding proceeds, and (ii) a persistent concentration on key semantic anchors, such as constraints, solution plans, and conclusions.

\begin{figure*}
  \centering
  \includegraphics[width=.99\textwidth]{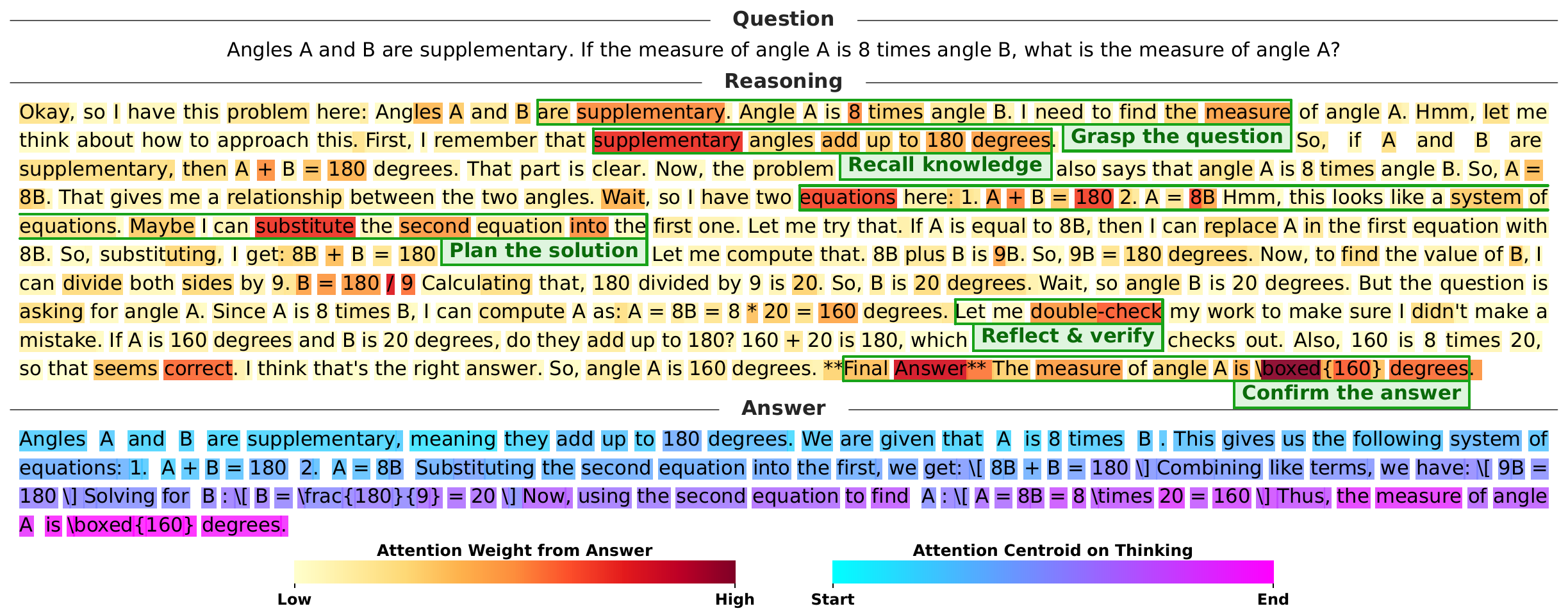}
  \caption{Visualization of benign self-reading in the middle-to-late layers of R1-Distill-Llama-8B. 
  The example is selected from the GSM8K \citep{gsm8k} benchmark.
  The answer panel colors tokens from blue to purple based on the attention centroid location to show the forward shift. 
  The centroid is defined as the weighted average center of an answer token’s attention distribution across reasoning tokens.
  The reasoning panel uses a gradient from yellow to red to indicate stronger attention accumulation from the answer tokens on pivotal semantic anchors.}
  \label{fig:example}
\end{figure*}

Figure \ref{fig:example} visualizes a typically benign self-reading pattern observed in a correct solution. 
In the answer panel, the shading shows that the reading focus of answer tokens moves forward along the reasoning trace: bluer tokens attend to earlier positions (e.g., ``Angles A and B are supplementary''), while purpler tokens attend later steps (e.g., ``\textbackslash boxed\{160\} degrees''). 
The reasoning panel highlights intense attention on key semantic steps (e.g., ``Recall knowledge'' and ``Plan the solution''). 
As the answer is decoded, the dependency centroid gradually shifts toward later reasoning positions. 
Simultaneously, attention repeatedly returns to key semantic anchors like reflections and conclusions.

\begin{figure*}
  \centering
  \includegraphics[width=.97\textwidth]{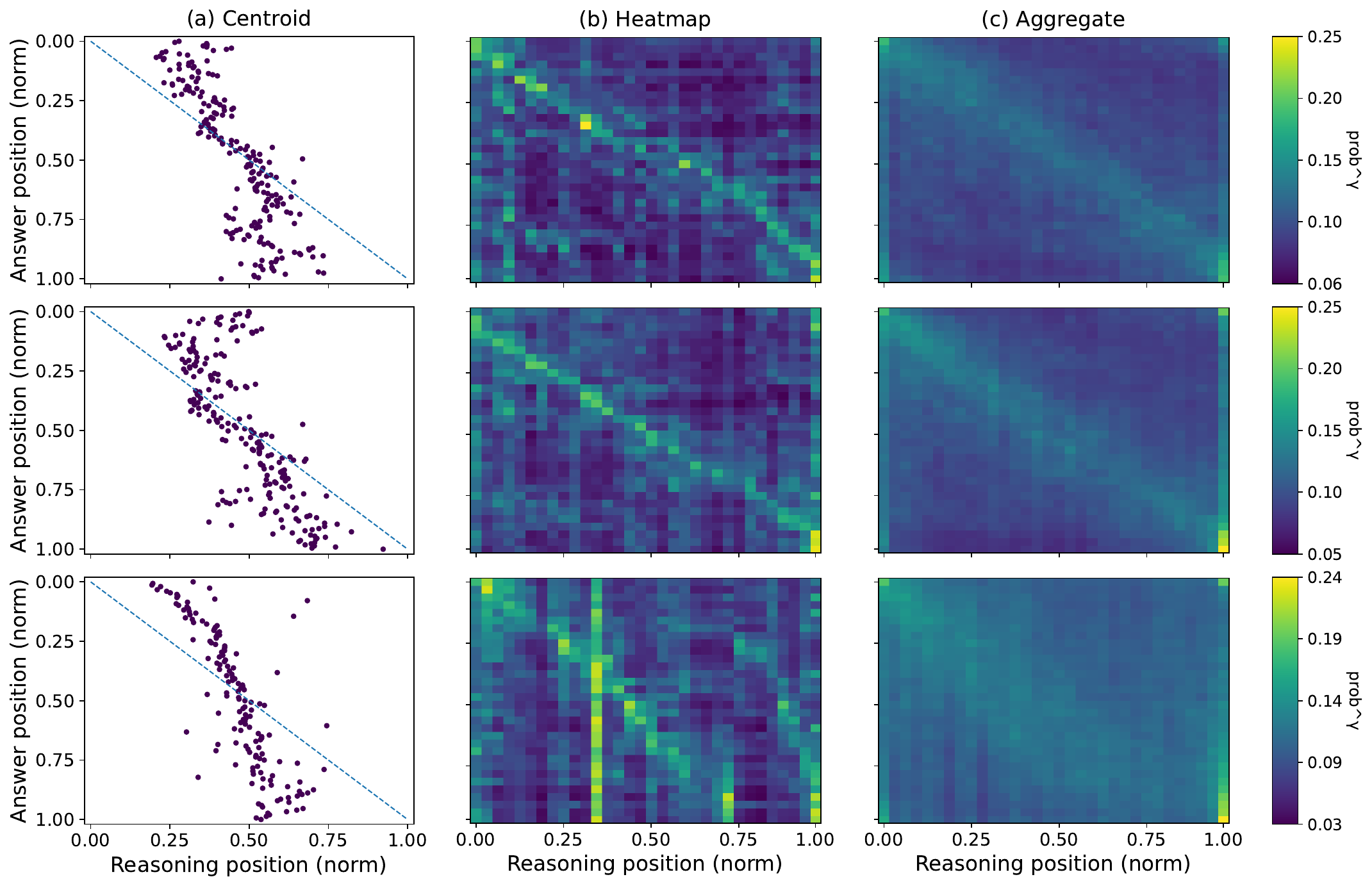}
  \caption{Attention visualization of benign self-reading on GSM8K. 
  Rows show R1-Distill-Llama-8B, R1-Distill-Qwen-7B, and Qwen3-4B-Thinking. Column (a) displays attention centroids, where each point represents the weighted average center of an answer token's attention distribution over the reasoning.
  Columns (a) and (b) use the same problem, while Column (c) shows the aggregated heatmaps averaged over 100 correct solutions per model.}
  \label{fig:analysis_vis}
\end{figure*}

These behaviors reveal that thinking LLMs strategically read their own reasoning traces during answer generation. 
We interpret this benign self-reading as a behavioral signature of internal certainty: by the answer stage, the model has committed to a particular solution path and relies on a few key reasoning steps, e.g., problem constraints and conclusions, as the evidence for answer generation. 
This interpretation aligns with classic cognitive theories \citep{NelsonNarens1990, Koriat1997} that cognition involves a meta-level that processes information derived from object-level execution based on internal states by \emph{control} and \emph{monitoring}. 
In this analogy, reasoning tokens implement object-level computation, while answer tokens function as a meta-level operation that reads and distills the reasoning trace to support generation. 
The progressive forward drift of the attention centroid reflects \emph{control}: the model advances its reading focus along a chosen branch instead of wandering. 
Meanwhile, the persistent focus on semantic anchors implements \emph{monitoring} by repeatedly revisiting the evidence that supports the answer tokens. 
Benign self-reading thus marks a stable, orderly, and internally confident state that typically coincides with correct solutions.
In contrast, disorganized attention in incorrect solutions signals cognitive uncertainty.
Section~\ref{subsec:uncertain} will show an example.

Building upon these insights, we propose a steering method driven by Self-Reading Quality (SRQ) scores to formalize and promote benign reading behaviors. 
SRQ measures the quality of a model’s self-reading from both geometric and semantic dimensions. 
The geometric dimension captures the process structure of control strategies, where metrics represent decisive execution along a solution path. 
The semantic dimension assesses the content grounding of monitoring quality by verifying whether the attended reasoning token is a critical step. 
By selecting samples with different SRQ scores, we construct steering vectors that guide LLMs toward benign self-reading and away from disorganized patterns associated with cognitive uncertainty. 
Our experiments on a range of thinking LLMs show that our method consistently boosts accuracy by steering the models toward a more decisive and grounded answer stage.

Our main contributions are outlined as follows:
\begin{itemize}
\item We identify the benign self-reading pattern during answer decoding on quantitative reasoning tasks. 
The pattern reflects the internal certainty and affects correctness.

\item We propose an SRQ-driven steering method. 
It guides the model toward more certain and well-organized self-reading patterns.

\item We evaluate our method on multiple quantitative reasoning benchmarks. 
Results show an accuracy improvement up to 2.6\% over the base LLMs, confirming the link between self-reading and answer correctness.
\end{itemize}

%====================%
\section{Related Work}

Activation engineering steers language model outputs by modifying intermediate activations during inference. 
\citet{caa} propose activation addition, an inference-time method that builds a steering vector from contrastive samples and adds it to hidden states to control attributes such as sentiment and toxicity.
\citet{ec} identify reasoning critical feed forward neurons based on activation differences in strong and weak reasoning traces.
\citet{honest} use steering vectors to generate safety steered honest alternatives, which helps LLM judges detect dishonesty reliably.
\citet{instruct} compute directions from activation differences with and without instructions and steer models to satisfy constraints on format length and required words, even with insufficient prompts.

Recent studies investigate activation engineering within thinking LLMs that separate reasoning and answering stages. 
\citet{understanding} extract steering vectors from DeepSeek-R1-Distill models to steer example-based verification and backtracking.
They find that these behaviors align with simple linear directions in activation space.
\citet{ASE} propose steered compression, extracting a vector that moves generation from verbose to concise reasoning and injecting it to reduce reasoning length while largely preserving accuracy.
\citet{sae} leverage the activation difference between symbolic mathematical generations and natural language generations to steer the model toward enhanced logical performance.
\citet{edit} identify a direction that controls reasoning length and edit projection weights of attention heads. 
As far as we know, these works have neither analyzed nor steered how answer tokens read the reasoning trace during answer decoding.

%====================%
\section{Self-Reading in Thinking LLMs}
\label{sec:pattern}
%Early research in cognitive psychology characterizes meta-cognition as an interaction between an \textit{object level}, which executes the primary task, and a \textit{meta level}, which processes available information via \textit{monitoring} and \textit{control} \citep{NelsonNarens1990, Koriat1997}.For Thinking LLMs, reasoning tokens constitute the object-level computation, while answer tokens perform meta-level operations by interpreting the generated trace.The meta-level processing of reasoning information does not merely determine the final output; it also serves as an indicator of the quality of the underlying object-level reasoning.

We conduct an analysis of the answer-to-reasoning attention mechanism on the GSM8K quantitative reasoning task \citep{gsm8k}. 
We utilize three frontier thinking LLMs: R1-Distill-Llama-8B \citep{r1}, R1-Distill-Qwen-7B \citep{r1}, and Qwen3-4B-Thinking \citep{qwen3}. 
We focus on the middle-to-late layers, which are regarded more semantically rich and stable \citep{m2,m1,from}. 

\begin{table}
  \centering
  {\small
  \begin{tabular}{lccccc}
    \toprule
    \textbf{Type} 
      & $C_{\text{SR}^+}$ 
      & $C_{\text{SR}^-}$ 
      & $I_{\text{SR}^+}$ 
      & $I_{\text{SR}^-}$ 
      & $Dis$ \\
    \midrule
    \textbf{Count} 
      & 159 
      & 12
      & 3
      & 23
      & 2$^+$\,+\,1$^-$ \\
    \bottomrule
  \end{tabular}}
  \caption{
    Human annotations on 200 randomly sampled GSM8K solutions. 
  }
  \label{tab:table1} 
\end{table}

%--------------------%
\subsection{Main Analysis}
We find that correct solutions exhibit a structured self-reading pattern during answer decoding, characterized by two distinct features:

\textbf{Forward-shifting attention centroid.}
First, the reading focus of answer tokens progressively shifts from earlier to later reasoning positions as decoding proceeds. 
To quantify this, we normalize the attention scores of each answer token over the thinking tokens and compute the weighted average position (centroid) of its focus. 
As visualized in Figure~\ref{fig:analysis_vis}(a), with positions normalized to $[0, 1]$, the resulting trajectory follows a clear diagonal path, indicating that the attention centroid moves synchronously with the generation of the answer. 
Figure~\ref{fig:analysis_vis}(b) also shows that the reading focus of answer tokens advances along the reasoning positions, forming a diagonal trend across models, with occasional parallel secondary bands.
The forward-shift of reading focus suggests that the model captures global information by following the logical flow of the reasoning trace. 
It indicates that the model deterministically tracks an effective solution path. We hypothesize that this sequential alignment implies that the reasoning content itself is logically coherent and the smooth synthesis of prior thoughts during the answer generation. See Appendix \ref{app:general_errors} for the contrasting visualizations of failure cases.

\textbf{Semantic anchor concentration.}
Beyond the global drift, the model relies on key semantic information in the reasoning trace to generate answer tokens. 
We observe that the answer tokens repeatedly concentrate on key \textit{semantic anchors}, such as problem constraints, solution plans, reflections, and final conclusions, revisiting a small set of crucial steps.
As shown in the heatmaps of Figure~\ref{fig:analysis_vis}(b), this behavior manifests as high-attention hotspots along the diagonal as well as bright off-diagonal anchors.
These local highlights underscore that the model strategically retrieves and reinforces key details rather than spreading attention uniformly. Together with the two features, the benign self-reading serves as an signature of the model’s \emph{internal certainty} when generating the answer.

To verify the universality of the benign self-reading patterns, we aggregate the attention maps from 100 correctly solved samples. 
In Figure~\ref{fig:analysis_vis}(c), the aggregated heatmaps maintain a distinct, soft diagonal ridge across all models. 
This confirms that the structured self-reading pattern is not an anecdotal but a stable and consistent behavior intrinsic to correct reasoning in thinking LLMs.

%--------------------%
\subsection{Benign Self-Reading and Correctness}
To validate the link between self-reading and accuracy, we conduct a human assessment on 200 solutions from R1-Distill-Llama-8B. 
We assign an example to one of four categories only when all three human annotators agree: correct with benign self-reading ($C_{\text{SR}^+}$), correct without benign self-reading ($C_{\text{SR}^-}$), incorrect with benign self-reading ($I_{\text{SR}^+}$), and incorrect without benign self-reading ($I_{\text{SR}^-}$). 
Examples with disagreement are grouped into a separate category ($Dis$).

We first annotate 200 randomly sampled solutions. 
Table~\ref{tab:table1} shows that benign self-reading is common in correct solutions but rare in incorrect ones. 
Since this random sampling is naturally imbalanced, we further annotate a balanced subset with equal numbers of correct and incorrect solutions. 
The result in Table~\ref{tab:balanced_annotation} confirms the same trend, indicating that benign self-reading is closely associated with correct and internally stable reasoning. 
Appendix~\ref{app:more_analysis} gives more analysis.

%Table~\ref{tab:table1} indicates that most correct solutions display benign self-reading, whereas incorrect solutions rarely do. Among the 173 correct solutions, approximately 92\% of the samples exhibit a clear benign self-reading pattern.

\begin{table}[t]
  \centering
  {\small
  \begin{tabular}{lccccc}
    \toprule
    \textbf{Type} 
      & $C_{\text{SR}^+}$ 
      & $C_{\text{SR}^-}$ 
      & $I_{\text{SR}^+}$ 
      & $I_{\text{SR}^-}$ 
      & $Dis$ \\
    \midrule
    \textbf{Count} 
      & 48 
      & 2
      & 4
      & 46 
      & 0\\
    \bottomrule
  \end{tabular}}
  \caption{
    Human annotations on a balanced subset of GSM8K samples, including 50 correct and 50 incorrect solutions.
  }
  \label{tab:balanced_annotation}
\end{table}

%====================%
\section{Self-Reading Steering Method}

Motivated by our analysis above, we propose a steering method driven by \emph{Self-Reading Quality} (SRQ) scores, which quantify how effectively answer tokens read the model’s reasoning traces.

%--------------------%
\subsection{From Self-Reading to Internal Certainty}
Early research \citep{NelsonNarens1990, Koriat1997} in cognitive psychology characterizes meta-cognition as an interaction between an \emph{object level} that executes the primary task and a \emph{meta level} that processes available information via \emph{monitoring} and \emph{control}. 
In thinking LLMs, reasoning tokens implement the object-level operation to reason, while answer tokens that read and summarize the generated reasoning trace acting as the meta-level operation. 

Benign self-reading in thinking LLMs aligns with this meta-cognitive theory through two features: 
(i) A forward-shifting reading focus acts as \emph{control}: the meta level utilizes object-level information by steadily advancing attention along a chosen solution path instead of wandering around irrelevant regions. 
(ii) High-attention anchors on key reasoning steps serve as \emph{monitoring}: the meta level inspects important steps of the reasoning to evaluate and support answer generation. 
This structured control and monitoring indicate  a natural signature of \emph{internal certainty}.

Accordingly, we decompose SRQ into a \emph{geometric} dimension that measures process control (how consistently attention progresses along the reasoning trace) and a \emph{semantic} dimension that measures content monitoring (whether attention concentrates on high-value semantic anchors and avoids misleading regions). 
We formalize the two dimensions below and show how the resulting SRQ scores enable reliable sample selection and effective steering.

%--------------------%
\subsection{Geometric Self-Reading Quality}
\label{sec:srq-geo}
The \emph{geometric} dimension measures whether answer tokens commit to a solution path and move steadily forward along the reasoning trace instead of wandering. 
Specifically, we use the \emph{global} and \emph{local} metrics to evaluate the forward diagonal trend of attention centroid movement.

To quantitatively track the reading focus of answer tokens over the reasoning trace, we define the \textit{attention centroid}. 
Let $P \in \mathbb{R}^{A \times T}$ be the answer-to-reasoning attention submatrix, with $A$ answer tokens and $T$ reasoning tokens. We form a row-normalized version $\tilde{P}_{ij} = \frac{P_{ij}}{\sum_{k=1}^{T} P_{ik} + \varepsilon}$, so each answer token defines a distribution over reasoning positions.
We index answer and reasoning positions by their normalized coordinates $x_i = \frac{i}{A}$ and $t_j = \frac{j}{T}$. %where $x_1$ is the first answer token
We define the attention centroid of answer token $i$ on the reasoning trace as $y_i = \sum_{j=1}^{T} t_j\, \tilde{P}_{ij}$.
The sequence $(x_i, y_i)$ tracks how the reading focus moves during answer decoding.

\textbf{Global forward-reading pattern.}
To capture the smooth forward drift of the centroid in benign self-reading, we define two global scores. 
The first is the \emph{Pearson correlation} between $x$ and $y$:
\begin{equation}
SRQ_{\mathrm{corr}} = \mathrm{corr}\left(x, y\right),
\end{equation}
where $x=(x_1,x_2\ldots,x_{A})$ and $y=(y_1,y_2\ldots,$ $y_{A})$. 
It achieves maximal when later answer tokens read later reasoning tokens. 

The second measures how closely the centroid trajectory follows the diagonal of the attention map:
\begin{equation}
SRQ_{\mathrm{diag}} = 1 - \sqrt{\frac{1}{A}\sum_{i=1}^{A} \left(y_i - x_i\right)^2}.
\end{equation}

\textbf{Local forward-reading consistency.} While the global scores capture the diagonal trend, benign self-reading also exhibits locally forward-moving focus. 
We measure local consistency using a sliding-window of size $w$ coverage score, which evaluates whether high-attention points exhibit a forward progression.
For each answer token, we extract its high-attention points by selecting the minimal set of reasoning positions with the largest $\tilde{P}_{ij}$ whose cumulative sum reaches $\beta$. 
Each point maps to the coordinates of normalized answer and reasoning positions. 
Then, for each of the $U$ overlapping windows, where $U=A-w+1$, we compute a \emph{Pearson correlation} $r_k$ between these coordinates. 
$\text{SRQ}_{\mathrm{local\_cover}}$, representing the fraction of windows exceeding a threshold $\tau$, is defined as:
\begin{equation}
SRQ_{\mathrm{local\_cover}} = \frac{1}{U}\sum_{u=1}^{U} \mathds{1}\left(r_k > \tau\right).
\end{equation}

%--------------------%
\subsection{Semantic Self-Reading Quality}
\label{sec:srq-sem}
On the \emph{semantic} dimension, answer attention is expected to repeatedly return to key semantic anchors rather than on noisy or incorrect ones. 
For supervision, we annotate each solution with span-level labels by LLM APIs (e.g., GPT-5 and Gemini-3-pro-preview) and project them to token indices. 
This yields a set $G$ of good reasoning tokens (e.g., constraints, correct intermediate steps), a set $B$ of bad reasoning tokens (e.g., incorrect computations, misleading branches), and a set $K_{\mathrm{ans}}$ of key answer tokens such as the final numerical results. 

\textbf{Semantic quality of reasoning focus.}
We first measure how the answer stage distributes its total attention flow over good and bad reasoning steps. 
Based on the raw matrix $P$, we compute a column sum $c_j = \sum_{i=1}^A P_{ij}$ for each reasoning token $j$ and select the top $\gamma T$ columns with the largest $c_j$ as a high-flow set $H$. 
The fraction on good ($G$) or bad ($B$) tokens is
\begin{equation}
SRQ_{\mathrm{think}}^{\pm} =
\frac{\sum_{j\in H\cap S} c_j}{\sum_{j\in H} c_j}, \ \ S \in \{G, B\}.
\end{equation}
%These metrics indicate whether the model focuses on correct derivations while avoiding wrong steps.

\textbf{Support quality of the answer.}
We measure whether key answer tokens attend to correct supporting steps. 
Given $K_{\mathrm{ans}}$, we define $SRQ_{\mathrm{ans}}^{\pm}$ as the average attention that they allocate to good or bad tokens:
\begin{equation}
SRQ_{\mathrm{ans}}^{\pm} = \frac{\sum\limits_{i\in K_{\mathrm{ans}}}\sum\limits_{j\in S} \tilde{P}_{ij}}{|K_{\mathrm{ans}}|}, \ \ S\in\{G, B\}.
\end{equation}

\textbf{Boundary emphasis on constraints and conclusions.}
Benign self-reading also concentrates attention on problem constraints and final conclusions. 
Using $P$, we define $m_{\mathrm{start}}$ and $m_{\mathrm{end}}$ as the attention sum on the first and last $\rho_{\mathrm{bd}}$ fraction of reasoning tokens. 
Instead of a uniform allocation giving $\rho_{\mathrm{bd}}$, we target mild enrichment with $\rho_{\mathrm{tar}}=\alpha\rho_{\mathrm{bd}}$ ($\alpha>1$). 
We score deviation from this target as follows:
\begin{equation}
SRQ_{\mathrm{start/end}} = \frac{m_{\mathrm{start/end}} - \rho_{\mathrm{tar}}}{1 - \rho_{\mathrm{tar}}}.
\end{equation}

%--------------------%
\subsection{SRQ-Based Sample Selection and Steering}
\label{sec:srq-selection}
With the geometric and semantic metrics, we build an integrated SRQ score per solution, and use it to select the contrastive sets and steering directions.

\textbf{From metrics to SRQ.}
For each solution $n$, we first rescale each geometric or semantic metric $SRQ^{(n)}_k$ to $s^{(n)}_k = \phi_k\left(SRQ^{(n)}_k\right) \in [0,1]$,
where $\phi_k$ is a monotone map fitting the empirical distribution on training solutions.
For the higher-is-better metrics (i.e., $SRQ_{\mathrm{corr}}$, $SRQ_{\mathrm{diag}}$, $SRQ_{\mathrm{local\_cover}}$, $SRQ_{\mathrm{corr}}$, $SRQ_{\mathrm{think}}^{+}$, and $SRQ_{\mathrm{ans}}^{+}$), large values are mapped close to 1; 
for the lower-is-better metrices ($SRQ_{\mathrm{think}}^{-}$ and $SRQ_{\mathrm{ans}}^{-}$), small values are mapped close to 1; and for the boundary metrics ($SRQ_{\mathrm{start}}$ and $SRQ_{\mathrm{end}}$), $\phi_k$ peaks near zero and decreases toward 0 for extreme concentration.
Then, we average scores within each group,
$\tilde{s}^{(n)}_{\alpha} = \frac{1}{|{K}_\alpha|}\sum_{k\in{K}_\alpha} s^{(n)}_k, \alpha \in \{\mathrm{geo}, \mathrm{sem}\}$,
and define the integrated SRQ score as follows:
\begin{equation}
\widetilde{SRQ}^{(n)} = \tilde{s}^{(n)}_{\mathrm{geo}} + \lambda_{\mathrm{sem}}\, \tilde{s}^{(n)}_{\mathrm{sem}}.
\end{equation}

\textbf{High and low-SRQ sample selection.}
%We use $SRQ^{(n)}$ to select solutions with clearly benign or disorganized self-reading.
We rank correct solutions $\mathcal{C}$ and incorrect solutions $\mathcal{I}$ by $\widetilde{SRQ}^{(n)}$.
We keep the top 80\% of $\mathcal{C}$ as $\mathcal{C}_{SRQ^+}$ and the bottom 80\% of $\mathcal{I}$ as $\mathcal{I}_{SRQ^-}$, yielding the contrastive groups differing in correctness and self-reading quality.

\textbf{Steering direction construction.}
Given $\mathcal{C}_{SRQ^+}$ and $\mathcal{I}_{SRQ^-}$, we construct benign self-reading steering directions using standard activation mechanisms.
We use the classic CAA \citep{caa} as an example.
At a chosen layer, we compute the mean token activations $\mu^{+,(s)}$ and $\mu^{-,(s)}$ for stage $s\in\{\mathrm{ans},\mathrm{reason}\}$ and define
\begin{equation}
v^{(s)} = \mu^{+,(s)} - \mu^{-,(s)}.
\end{equation}

At inference, we apply $v^{(\mathrm{ans})}$ to answer-token activations and $v^{(\mathrm{reason})}$ during reasoning. The SRQ scores can also be combined with other mainstream steering mechanisms such as \citep{conce,pca}. 

\begin{table*}[t]
\centering
\resizebox{\textwidth}{!}{
\begin{tabular}{clccccccc}
\toprule
\multirow{2}{*}{\textbf{Dataset}} &
\multirow{2}{*}{\textbf{Thinking LLM}} &
\multirow{2}{*}{\textbf{Base}} &
\multicolumn{3}{c}{\textbf{Steering Mechanism}} &
\multicolumn{3}{c}{\textbf{+\,Self-Reading Steering}} \\
\cmidrule(lr){4-6}\cmidrule(lr){7-9}
& & & \textbf{CAA} & \textbf{Conceptor} & \textbf{PCA-CAA}
  & \textbf{+\,CAA}
  & \textbf{+\,Conceptor}
  & \textbf{+\,PCA-CAA} \\
\midrule

\multirow{3}{*}{GSM8K} 
& R1-Distill-Qwen-7B & 87.8 & 88.4 & 88.6 & 89.1 & 89.5\,(1.7$\uparrow$) & 89.6\,(1.8$\uparrow$) & \textbf{89.9\,(2.1$\uparrow$)} \\
& R1-Distill-Llama-8B & 85.8 & 86.7 & 87.1 & 87.5 & 88.1\,(2.3$\uparrow$) & 88.2\,(2.4$\uparrow$) & \textbf{88.4\,(2.6$\uparrow$)} \\
& Qwen3-4B-Thinking & 91.9 & 92.2 & 92.6 & 92.4 & 93.0\,(1.1$\uparrow$) & \textbf{93.3\,(1.4$\uparrow$)} & 93.1\,(1.2$\uparrow$) \\
\midrule

\multirow{3}{*}{MATH500} 
& R1-Distill-Qwen-7B  & 88.6 & 89.2 & 89.6 & 89.8 & 90.0\,(1.4$\uparrow$) & \textbf{90.4\,(1.8$\uparrow$)} & \textbf{90.4\,(1.8$\uparrow$)} \\
& R1-Distill-Llama-8B & 85.6 & 86.0 & 86.5 & 87.1 & 87.0\,(1.4$\uparrow$) & 87.4\,(1.8$\uparrow$) & \textbf{87.9\,(2.3$\uparrow$)} \\
& Qwen3-4B-Thinking & 94.2 & 94.2 & 94.6 & 94.4 & 94.8\,(0.6$\uparrow$) & \textbf{95.6\,(1.4$\uparrow$)} & 95.2\,(1.0$\uparrow$) \\
\midrule

\multirow{3}{*}{SVAMP}
& R1-Distill-Qwen-7B & 90.3 & 91.1 & 91.6 & 92.0 & 92.2\,(1.9$\uparrow$) & 92.5\,(2.2$\uparrow$) & \textbf{92.7\,(2.4$\uparrow$)} \\
& R1-Distill-Llama-8B & 90.7 & 91.6 & 92.0 & 92.4 & 92.3\,(1.6$\uparrow$) & 92.6\,(1.9$\uparrow$) & \textbf{93.2\,(2.5$\uparrow$)} \\
& Qwen3-4B-Thinking & 94.5 & 95.2 & 95.4 & 95.5 & 95.8\,(1.3$\uparrow$) & 96.1\,(1.6$\uparrow$) & \textbf{96.5\,(2.0$\uparrow$)} \\
\bottomrule
\end{tabular}}
\caption{Accuracy on GSM8K, MATH500, and SVAMP. The gains of our method over base LLMs are also shown.}
\label{tab:main_results}
\end{table*}

%====================%
\section{Experiments and Results}
\label{sec:exp}

\subsection{Experimental Setup}

\textbf{Models.} We employ three frontier thinking LLMs with explicit reasoning traces and short answer stages: R1-Distill-Qwen-7B, R1-Distill-Llama-8B, and Qwen3-4B-Thinking. 
Decoding for all models uses temperature $=0.65$ and top-$p = 0.95$.

\textbf{Datasets.} Our evaluation uses three quantitative reasoning benchmarks: GSM8K~\citep{gsm8k}, MATH500 \citep{math500}, and SVAMP \citep{svamp}. 
We construct steering vectors on the training sets of these datasets, where SVAMP is particularly considered for transfer evaluation with steering vectors from GSM8K. 
All results are reported using accuracy.

\textbf{Steering mechanisms.} To assess the compatibility with different steering methods, we instantiate classic CAA~\citep{caa}, conceptor steering~\citep{conce}, and recent PCA-CAA~\citep{pca}. 
We follow the standard practice to construct activation directions. 
See Appendix~\ref{app:ex} for more details about dataset usage, layer selection, and steering construction.

%--------------------%
\subsection{Main Results}
We first test whether self-reading signals can serve as a universal supervision source compatible with various steering mechanisms. 
Table~\ref{tab:main_results} reports the accuracy on GSM8K, MATH500, and SVAMP. 

Across all datasets and models, our self-reading steering improves the base models and strengthens existing steering mechanisms. 
The gains appear consistently for CAA, Conceptor, and PCA-CAA, indicating that SRQ is a generic layer steering the model toward more certain, stable internal states across different mechanisms.

On GSM8K, all three steering mechanisms already bring clear improvements over the non-steered base models. 
For example, R1-Distill-Qwen-7B rises from 87.8 to 89.1 with PCA-CAA. 
With our self-reading steering, the accuracy climbs further to 89.9. 
Qwen3-4B-Thinking is a competing base model with 91.9 accuracy, yet self-reading steering still lifts it to 93.3 with Conceptor. 
These results indicate that our method yields additive gains by reinforcing a reading mode linked to higher internal certainty, thereby stabilizing LLMs' cognitive state and improving performance.

MATH500 is a more challenging benchmark with longer solutions, and we see the same trend. 
Even with long, noisy reasoning traces, guiding the model toward a benign self-reading stabilizes its internal control and monitoring. 
This facilitates a more decisive process for noisy reasoning contents during the answer stage.

SVAMP assesses how well the learned steering vectors transfer across datasets. 
Again, self-reading steering yields improvements for all models and mechanisms. 
The successful transfer to SVAMP confirms that our method targets a general, intrinsic reading strategy rather than overfitting to GSM8K.

\begin{table}
\centering
\resizebox{\columnwidth}{!}{%
\begin{tabular}{lccccc}
\toprule
\multirow{2}{*}{\textbf{LLM}} &
\multicolumn{3}{c}{\textbf{PCA-CAA\,+}} &
\multicolumn{2}{c}{\textbf{CAA\,+}} \\
\cmidrule(lr){2-4}\cmidrule(lr){5-6}
& \textbf{RePE} & \textbf{SAE-free} & \textbf{Ours} & \textbf{SEAL} & \textbf{Ours} \\
\midrule
R1-Qwen-7B  & 89.5 & 89.6 & \textbf{89.9} & 88.8 & \textbf{89.5} \\
R1-Llama-8B & 87.8 & 88.1 & \textbf{88.4} & 87.1 & \textbf{88.1} \\
\bottomrule
\end{tabular}
}
\caption{Comparison of steering methods for improving the reasoning trace on GSM8K. 
The accuracy of the two base models is 87.8 and 85.8.}
\label{tab:reasoning_opt_methods}
\end{table}

%--------------------%
\subsection{Comparison to Other Steering Methods}
To validate the effectiveness of self-reading, we compare it with three reasoning-focused steering methods that derive intervention directions from different properties of the reasoning trace: 
(i) RepE \citep{pca} steers the model away from chaotic and off-topic behavior toward clearer and more focused reasoning. 
(ii) SAE-free \citep{sae} analyzes reasoning traces and extracts directions that emphasize mathematical and logical structure while suppressing generic narrative content. 
(iii) SEAL \citep{seal} promotes concise and efficient reasoning by preferring execution-like representations and reducing reflection and transition. 
Following the literature, RepE and SAE-free use PCA-CAA, while SEAL uses CAA.

\begin{table}
\centering
\resizebox{\columnwidth}{!}{
\begin{tabular}{clccc} 
\toprule
\textbf{Dataset} & \textbf{LLM} & \textbf{Base} & \textbf{CAA} & \textbf{+\,CAA} \\
\midrule
\multirow{3}{*}{SciQ} & R1-Distill-Qwen-7B  & 85.9 & 86.4 & \textbf{86.7} \\
& R1-Distill-Llama-8B & 87.3 & 87.7 & \textbf{88.3} \\
& Qwen3-4B-Thinking & 93.9 & 94.2 & \textbf{94.6} \\
\midrule
\multirow{3}{*}{\makecell{AIME\\24--25}} & R1-Distill-Qwen-7B & 46.7 & 48.3 & \textbf{51.7} \\
& R1-Distill-Llama-8B & 41.7 & 45.0 & \textbf{48.3} \\
& Qwen3-4B-Thinking & 76.7 & 78.3 & \textbf{80.0} \\
\bottomrule
\end{tabular}}
\caption{Generalization to other quantitative reasoning tasks: SciQ and AIME24--25.}
\label{tab:gen_sciq_aime}
\end{table}

\begin{table}
\centering
\small
\resizebox{\columnwidth}{!}{
\begin{tabular}{lcccc}   
\toprule
\textbf{LLM} & \textbf{Base} & \textbf{-- Geom.} & \textbf{-- Sem.} & \textbf{Full} \\
\midrule
R1-Distill-Llama-8B & 85.8 & 87.3 & 87.6 & \textbf{88.1} \\
Qwen3-4B-Thinking & 91.9&92.5&92.7&\textbf{93.0}\\
\bottomrule
\end{tabular}
}
\caption{Ablation results of R1-Distill-Llama-8B and Qwen3-4B-Thinking on GSM8K. All variants use CAA.}
\label{tab:ablation_dsLlama}
\end{table}

Table~\ref{tab:reasoning_opt_methods} presents the accuracy comparison on GSM8K with R1-Distill-Qwen-7B and R1-Distill-Llama-8B. 
All steering methods improve the non-steered base models. 
Based upon the PCA-CAA mechanism, our self-reading steering reaches 89.9 and 88.4, surpassing both RepE and SAE-free. 
Similarly, with the CAA mechanism, our method outperforms SEAL. 
The consistency of improvements highlights the robustness of our method.
These results indicate that steering the model toward internally certain states characterized by benign self-reading patterns is a meaningful and effective objective.
The self-reading signals facilitate the selection of higher-quality samples for intervention, thereby leading to better performance.

\begin{figure}
\centering
\includegraphics[width=\linewidth]{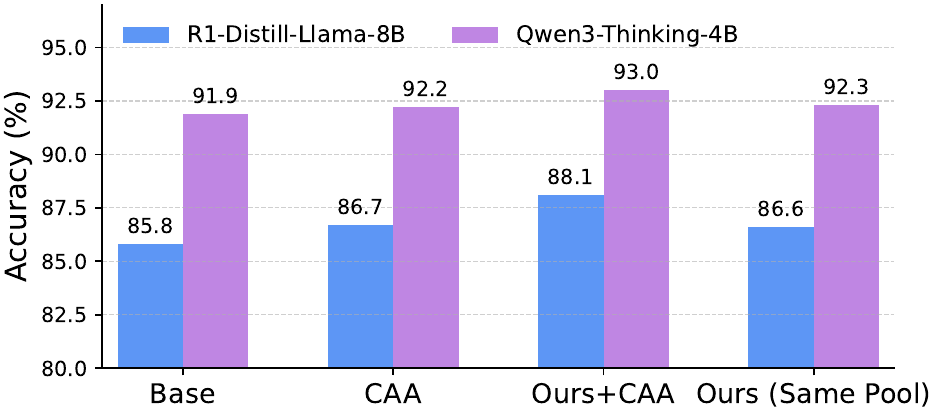}
\caption{Comparison on GSM8K with CAA. Instead of using a larger candidate pool to obtain 700 pairs, the ``Same Pool'' variant draws from the same pool sampled by CAA and filters it to retain a smaller set.}
\label{fig:GSM8K_pair_selection_fairness}
\end{figure}

%--------------------%
\subsection{Generalization to Broader Tasks}
To further explore the generalization across quantitative reasoning tasks, we evaluate our self-reading steering method on SciQ (Scientific Question Answering) \citep{sciq} and AIME24–25 \citep{aime_official} (American Invitational Mathematics Examination). 
These datasets require more complex reasoning chains \citep{WOS:001269604100009}.
We construct vectors from SciQ's training split and reuse MATH500 vectors for AIME. 
As shown in Table~\ref{tab:gen_sciq_aime}, our method consistently improves the base models and the CAA mechanism. 
Particularly, on the very difficult AIME24--25, the improvement reaches up to 6.6 points over the base model R1-Distill-Llama-8B, which corresponds to four additional competition problems solved. 
The success on these tasks confirms that our method filters a general signal of internal certainty.

%--------------------%
\subsection{Ablation Study on SRQ Dimensions}
\label{subsec:ablation_srq}
We conduct an ablation study to investigate the individual contributions of the geometric and semantic dimensions of SRQ scores. 
As shown in Table \ref{tab:ablation_dsLlama}, removing either dimension reduces performance, with the geometric component having a stronger impact on steering effectiveness. 
Through manual inspection of samples, we find that the good geometric features corresponding to forward-shifting attention may have already captured the key steps in these traces.

%--------------------%
\subsection{Analysis of Candidate Set Fairness}
\label{subsec:ablation_fair}
To isolate the benefit of SRQ selection from the potential impact of a larger candidate pool, we conduct a ``Same Pool" analysis. 
We apply SRQ scores to the pairs of correct and incorrect traces used by the standard steering mechanism and filter them to retain a subset of pairs. 
As presented in Figure~\ref{fig:GSM8K_pair_selection_fairness}, the ``Same Pool'' variant remains competitive with or slightly outperforms the baseline while using fewer samples to construct the steering vectors. 
This verifies that activation steering vectors extracted from the smaller subset of samples by SRQ remain highly effective.

%--------------------%
\subsection{Case Study of Uncertain Self-Reading}
\label{subsec:uncertain}
\begin{figure}
  \includegraphics[width=\columnwidth]{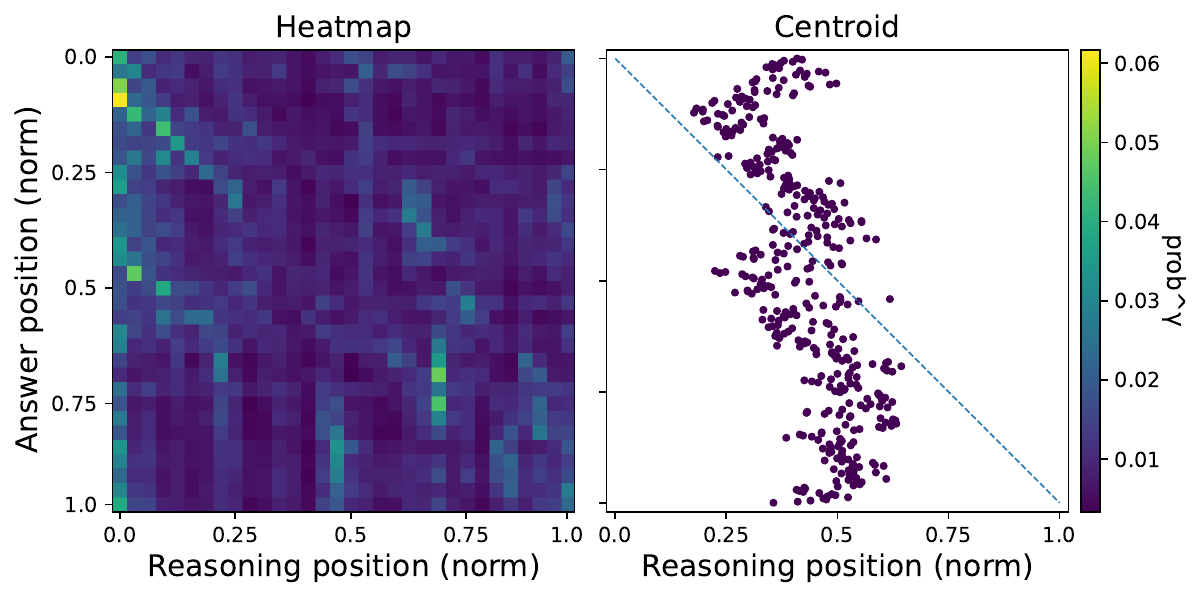}
  \caption{Visualization of a special self-reading error in R1-Distill-Llama-8B.}
  \label{fig:error}
\end{figure}

Incorrect answers often exhibit irregular attention without a consistent organization. 
Figure~\ref{fig:error} highlights a special counter case where the reasoning trace reaches the correct result but is long, cluttered, and filled with reflective ``waits'', yet the answer stage is wrong. 

The heatmap shows dark colors without clear oblique bands or anchor columns, indicating broadly distributed attention over the reasoning trace. Isolated bright spots near the problem statement remain inconsistent with the benign pattern.
The centroid plot shows that rather than a smooth forward drift, the centroids form a nearly vertical broken line that oscillates within the same region.
This suggests a failure of both control and monitoring: the model does not maintain a certain solution branch or lock onto semantic anchors during answer decoding. 
The model wanders within a noisy trace and fails to consolidate information for a correct prediction. 
More attention visualizations of general failure cases are detailed in Appendix \ref{app:general_errors}.

%====================%
\subsection{Analysis of Benign Self-reading and Cognitive Certainty}

We hypothesize that facilitating benign reading patterns induces a state of higher cognitive certainty, thereby enhancing reasoning performance.
To verify this, we compare the model's inference confidence with and without the intervention of SRQ activation vectors.
We quantify the model's overall certainty by computing the mean confidence over the complete generation process. 
The confidence is defined as follows:
\begin{align}
Conf = \left( \prod_{i=1}^{n} \max_{a_t \in V} p(a_t) \right)^{\frac{1}{n}},
\end{align}
where $p(a_t)=\mathrm{Pr}[a_t \,|\, H, I, a_{<t}]$ calculates the model's predicted probability for the answer token $a_t$. 
Here, $H$ denotes the context including the input prompt and generated thoughts, $I$ is the instruction to elicit the answer, $n$ is the number of decoding steps, $V$ is the vocabulary, and $a_{<t}$ denotes all previously generated tokens \citep{yiliu25, zhixindu}.

As presented in Table~\ref{tab:confidence}, guiding the model with the activation vector leads to consistent improvements in both accuracy and confidence, confirming that our method effectively boosts the model's internal certainty.

\begin{table}
\centering
\small
\begin{tabular}{lcccc}
\toprule
\multirow{2}{*}{\textbf{LLM}} & \multicolumn{2}{c}{\textbf{Base}} & \multicolumn{2}{c}{\textbf{Ours\,+\,CAA}} \\
\cmidrule(lr){2-3} \cmidrule(lr){4-5} & \textbf{Acc} & \textbf{Conf} & \textbf{Acc} & \textbf{Conf} \\
\midrule
R1-Distill-Llama-8B & 85.8 & 82.8\% & 88.1 & 84.1\% \\
\bottomrule
\end{tabular}
\caption{Comparison of accuracy and confidence on GSM8K.}
\label{tab:confidence}
\end{table}

%--------------------%
\subsection{Scaling Analysis}
\label{subsec:scaling}
We extend our evaluation to larger models to verify whether the effectiveness of our approach persists as model capacity increases. 
Table~\ref{tab:scale_7b_14b} lists that our self-reading steering consistently yields improvements on both the 7B and 14B models. 
Although the baseline performance of R1-Distill-Qwen-14B is already strong, our method still brings modest gains. 
These results demonstrate that the signals captured by SRQ remain robust and effective across different model scales.

%抑制反思的讨论，放两个统计的数据，一个是反转（最初错误但后来自我纠正），另一个是单纯关于reflect的统计；
%一个是对生成长度的影响 放附录 ok
%挪一个，把置信度那个挪上来 ok
%数据失衡 
%对失败案例多解释几句
%拼写错误要改 ok
%附录里api说清楚，然后数据筛选多用几句说清楚，ok

%====================%
\section{Conclusion}
We study how answer tokens read the reasoning trace in thinking LLMs for quantitative reasoning, and discover a benign self-reading pattern with forward-shifting attention centroids and repeated focus on semantic anchors that is strongly associated with model certainty. 
We propose a steering method driven by SRQ scores combining geometric and semantic metrics, which promotes certain internal states and well-organized self-reading at inference time. 
Our experiments show consistent accuracy gain, suggesting self-reading is an effective and generalizable signal, and SRQ is likely to remain useful as steering methods evolve.

\begin{table}
\centering
\resizebox{\columnwidth}{!}{
\begin{tabular}{clccc}
\toprule
\textbf{Dataset} & \textbf{LLM} & \textbf{Base} & \textbf{CAA} & \textbf{+\,CAA} \\
\midrule
\multirow{2}{*}{GSM8K} & R1-Distill-Qwen-7B & 87.8 & 88.4 & 89.5 \\
& R1-Distill-Qwen-14B& 94.6 & 94.9 & \textbf{95.2} \\
\midrule
\multirow{2}{*}{MATH500} & R1-Distill-Qwen-7B & 88.6 & 89.2 & 90.0 \\
& R1-Distill-Qwen-14B& 93.2 & 93.6 & \textbf{93.9} \\
\bottomrule
\end{tabular}}
\caption{Scaling results on GSM8K and MATH500.}
\label{tab:scale_7b_14b}
\end{table}

\section*{Acknowledgments}
This work was supported by the National Natural Science Foundation of China (No. 62272219) and the CCF-1688 Yuanbao Cooperation Fund (No. CCF-Alibaba2025001).

%====================%
% 不占页数限制
\section*{Ethical Considerations}
The datasets and benchmarks used in this work are publicly available and distributed under permissive licenses, and we follow their stated terms of use. 
For result analysis, we recruited graduate students to perform manual annotations, and informed consent was obtained from all participants.
Apart from these annotations, our work does not involve collecting new human data, personal or sensitive information, or interventions in real-world user-facing systems.
Nevertheless, as with any work on LLMs, our methods could inherit biases or limitations present in the underlying models or datasets, and results should be interpreted with this in mind.

%====================%
\section*{Limitations}
First, our study focuses on thinking LLMs that expose an explicit reasoning trace followed by a dedicated answer stage, and our SRQ metrics are defined over answer-to-reasoning attention and intermediate activations. 
This limits direct applicability to open-source thinking LLMs where such internal signals are accessible, and the best layer choices may vary across architectures.

Second, semantic SRQ relies on span-level annotations produced by external LLMs, which introduces additional cost and potential label noise. 
Developing cheaper and fully automatic proxies remains a future work. 
Also, the benefits of steering the answer stage are bounded by the quality of the available evidence in the reasoning trace. 
If the trace is fundamentally incorrect or lacks key steps, self-reading alone may have limited improvement headroom, motivating tighter coupling between reasoning-stage and answer-stage interventions.

Finally, our analysis is centered on quantitative reasoning, where the benign self-reading pattern appears most consistently, especially on math-style problems. 
For other domains, it remains unclear whether they exhibit the same or different self-reading patterns. 
Understanding how domain characteristics relate to self-reading behaviors is an important direction of future work.

%\bibliographystyle{custom}
\bibliography{custom}

@inproceedings {thinking,
  author       = { Tianhao Wu and Janice Lan and Weizhe Yuan and Jiantao Jiao and Jason E. Weston and Sainbayar Sukhbaatar },
  title        = { Thinking {LLMs}: General Instruction Following with Thought Generation },
  booktitle    = { Forty-second International Conference on Machine Learning, {ICML}
                  2025, Vancouver, BC, Canada, July 13-19, 2025 },
  year         = 2025,
  publisher    = { OpenReview.net },
  url          = { https://openreview.net/forum?id=z6SrgYCdey },
  bibsource    = { dblp computer science bibliography, https://dblp.org }
}

@article {r1,
  author       = { DeepSeek{-}AI },
  title        = { {DeepSeek-R1}: Incentivizing Reasoning Capability in {LLMs} via Reinforcement Learning },
  journal      = { CoRR },
  year         = 2025,
  url          = { https://doi.org/10.48550/arXiv.2501.12948 },
  doi          = { 10.48550/ARXIV.2501.12948 },
  eprinttype   = { arXiv },
  eprint       = { 2501.12948 },
  bibsource    = { dblp computer science bibliography, https://dblp.org }
}

@article {seal,
  author       = { Runjin Chen and Zhenyu Zhang and Junyuan Hong and Souvik Kundu and Zhangyang Wang },
  title        = { {SEAL}: Steerable Reasoning Calibration of Large Language Models for Free },
  journal      = { CoRR },
  year         = 2025,
  url          = { https://doi.org/10.48550/arXiv.2504.07986 },
  doi          = { 10.48550/ARXIV.2504.07986 },
  eprinttype   = { arXiv },
  eprint       = { 2504.07986 },
  bibsource    = { dblp computer science bibliography, https://dblp.org }
}

@inproceedings {instruct,
  author       = { Alessandro Stolfo and Vidhisha Balachandran and Safoora Yousefi and Eric Horvitz and Besmira Nushi },
  title        = { Improving Instruction-Following in Language Models through Activation Steering },
  booktitle    = { The Thirteenth International Conference on Learning Representations,
                  {ICLR} 2025, Singapore, April 24-28, 2025 },
  year         = 2025,
  publisher    = { OpenReview.net },
  url          = { https://openreview.net/forum?id=wozhdnRCtw },
  bibsource    = { dblp computer science bibliography, https://dblp.org }
}

@article {honest,
  author       = { Leon Eshuijs and Archie Chaudhury and Alan McBeth and Ethan Nguyen },
  title        = { But what is your honest answer? {Aiding} {LLM}-judges with honest alternatives using steering vectors },
  journal      = { CoRR },
  year         = 2025,
  url          = { https://doi.org/10.48550/arXiv.2505.17760 },
  doi          = { 10.48550/ARXIV.2505.17760 },
  eprinttype   = { arXiv },
  eprint       = { 2505.17760 },
  bibsource    = { dblp computer science bibliography, https://dblp.org }
}

@article {understanding,
  author       = { Constantin Venhoff and Iv{\'{a}}n Arcuschin and Philip Torr and Arthur Conmy and Neel Nanda },
  title        = { Understanding Reasoning in Thinking Language Models via Steering Vectors },
  journal      = { CoRR },
  year         = 2025,
  url          = { https://doi.org/10.48550/arXiv.2506.18167 },
  doi          = { 10.48550/ARXIV.2506.18167 },
  eprinttype   = { arXiv },
  eprint       = { 2506.18167 },
  bibsource    = { dblp computer science bibliography, https://dblp.org }
}

@article {ASE,
  author       = { Seyedarmin Azizi and Erfan Baghaei Potraghloo and Massoud Pedram },
  title        = { Activation Steering for Chain-of-Thought Compression },
  journal      = { CoRR },
  year         = 2025,
  url          = { https://doi.org/10.48550/arXiv.2507.04742 },
  doi          = { 10.48550/ARXIV.2507.04742 },
  eprinttype   = { arXiv },
  eprint       = { 2507.04742 },
  bibsource    = { dblp computer science bibliography, https://dblp.org }
}

@article {long_cot,
  author       = { Zekai Zhao and Qi Liu and Kun Zhou and Zihan Liu and Yifei Shao and Zhiting Hu and Biwei Huang },
  title        = { Activation Control for Efficiently Eliciting Long Chain-of-thought Ability of Language Models },
  journal      = { CoRR },
  year         = 2025,
  url          = { https://doi.org/10.48550/arXiv.2505.17697 },
  doi          = { 10.48550/ARXIV.2505.17697 },
  eprinttype   = { arXiv },
  eprint       = { 2505.17697 },
  bibsource    = { dblp computer science bibliography, https://dblp.org }
}

@article {from,
  author       = { Jue Zhang and Qingwei Lin and Saravan Rajmohan and Dongmei Zhang },
  title        = { From Reasoning to Answer: Empirical, Attention-Based and Mechanistic Insights into Distilled {DeepSeek R1} Models },
  journal      = { CoRR },
  year         = 2025,
  url          = { https://doi.org/10.48550/arXiv.2509.23676 },
  doi          = { 10.48550/ARXIV.2509.23676 },
  eprinttype   = { arXiv },
  eprint       = { 2509.23676 },
  bibsource    = { dblp computer science bibliography, https://dblp.org }
}

@article {Koriat1997,
  author       = { Asher Koriat },
  title        = { Monitoring one's own knowledge during study: A cue-utilization approach to judgments of learning. },
  journal      = { Journal of experimental psychology: General },
  year         = 1997,
  pages        = { 349 },
  volume       = 126,
  number       = 4,
  publisher    = { American Psychological Association }
}

@incollection {NelsonNarens1990,
  author       = { Thomas O Nelson },
  title        = { Metamemory: A theoretical framework and new findings },
  booktitle    = { Psychology of learning and motivation },
  publisher    = { Elsevier },
  year         = 1990,
  volume       = 26,
  pages        = { 125--173 }
}

@inproceedings {m1,
  author       = { Oscar Skean and Md Rifat Arefin and Dan Zhao and Niket Patel and Jalal Naghiyev and Yann LeCun and Ravid Shwartz{-}Ziv },
  title        = { Layer by Layer: Uncovering Hidden Representations in Language Models },
  booktitle    = { Forty-second International Conference on Machine Learning, {ICML}
                  2025, Vancouver, BC, Canada, July 13-19, 2025 },
  year         = 2025,
  publisher    = { OpenReview.net },
  url          = { https://openreview.net/forum?id=WGXb7UdvTX },
  bibsource    = { dblp computer science bibliography, https://dblp.org }
}

@inproceedings {m2,
  author       = { Anna Langedijk and Hosein Mohebbi and Gabriele Sarti and Willem H. Zuidema and Jaap Jumelet },
  title        = { {DecoderLens}: Layerwise Interpretation of Encoder-Decoder Transformers },
  booktitle    = { Findings of the Association for Computational Linguistics: {NAACL}
                  2024, Mexico City, Mexico, June 16-21, 2024 },
  year         = 2024,
  pages        = { 4764--4780 },
  publisher    = { Association for Computational Linguistics },
  editor       = { Kevin Duh and
                  Helena G{\'{o}}mez{-}Adorno and
                  Steven Bethard },
  url          = { https://doi.org/10.18653/v1/2024.findings-naacl.296 },
  doi          = { 10.18653/V1/2024.FINDINGS-NAACL.296 },
  bibsource    = { dblp computer science bibliography, https://dblp.org }
}

@inproceedings {m3,
  author       = { Ala N. Tak and Amin Banayeeanzade and Anahita Bolourani and Mina Kian and Robin Jia and Jonathan Gratch },
  title        = { Mechanistic Interpretability of Emotion Inference in Large Language Models },
  booktitle    = { Findings of the Association for Computational Linguistics, {ACL} 2025,
                  Vienna, Austria, July 27 - August 1, 2025 },
  year         = 2025,
  pages        = { 13090--13120 },
  publisher    = { Association for Computational Linguistics },
  editor       = { Wanxiang Che and
                  Joyce Nabende and
                  Ekaterina Shutova and
                  Mohammad Taher Pilehvar },
  url          = { https://aclanthology.org/2025.findings-acl.679/ },
  bibsource    = { dblp computer science bibliography, https://dblp.org }
}

@inproceedings {m4,
  author       = { Yu Zhao and Alessio Devoto and Giwon Hong and Xiaotang Du and Aryo Pradipta Gema and Hongru Wang and Xuanli He and Kam{-}Fai Wong and Pasquale Minervini },
  title        = { Steering Knowledge Selection Behaviours in {LLMs} via {SAE}-Based Representation Engineering },
  booktitle    = { Proceedings of the 2025 Conference of the Nations of the Americas
                  Chapter of the Association for Computational Linguistics: Human Language
                  Technologies, {NAACL} 2025 - Volume 1: Long Papers, Albuquerque, New
                  Mexico, USA, April 29 - May 4, 2025 },
  year         = 2025,
  pages        = { 5117--5136 },
  publisher    = { Association for Computational Linguistics },
  editor       = { Luis Chiruzzo and
                  Alan Ritter and
                  Lu Wang },
  url          = { https://doi.org/10.18653/v1/2025.naacl-long.264 },
  doi          = { 10.18653/V1/2025.NAACL-LONG.264 },
  bibsource    = { dblp computer science bibliography, https://dblp.org }
}

@article {Ali2025EntropyLensTI,
  author       = { Riccardo Ali and Francesco Caso and Christopher Irwin and Pietro Li{\`{o}} },
  title        = { {Entropy-Lens}: The Information Signature of Transformer Computations },
  journal      = { CoRR },
  year         = 2025,
  url          = { https://doi.org/10.48550/arXiv.2502.16570 },
  doi          = { 10.48550/ARXIV.2502.16570 },
  eprinttype   = { arXiv },
  eprint       = { 2502.16570 },
  bibsource    = { dblp computer science bibliography, https://dblp.org }
}

@article {qwen3,
  author       = { An Yang and Anfeng Li and Baosong Yang and Beichen Zhang and Binyuan Hui and Bo Zheng and Bowen Yu and Chang Gao and Chengen Huang and Chenxu Lv and Chujie Zheng and Dayiheng Liu and Fan Zhou and Fei Huang and Feng Hu and Hao Ge and Haoran Wei and Huan Lin and Jialong Tang and Jian Yang and Jianhong Tu and Jianwei Zhang and Jian Yang and Jiaxi Yang and Jingren Zhou and Junyang Lin and Kai Dang and Keqin Bao and Kexin Yang and Le Yu and Lianghao Deng and Mei Li and Mingfeng Xue and Mingze Li and Pei Zhang and Peng Wang and Qin Zhu and Rui Men and Ruize Gao and Shixuan Liu and Shuang Luo and Tianhao Li and Tianyi Tang and Wenbiao Yin and Xingzhang Ren and Xinyu Wang and Xinyu Zhang and Xuancheng Ren and Yang Fan and Yang Su and Yichang Zhang and Yinger Zhang and Yu Wan and Yuqiong Liu and Zekun Wang and Zeyu Cui and Zhenru Zhang and Zhipeng Zhou and Zihan Qiu },
  title        = { Qwen3 Technical Report },
  journal      = { CoRR },
  year         = 2025,
  url          = { https://doi.org/10.48550/arXiv.2505.09388 },
  doi          = { 10.48550/ARXIV.2505.09388 },
  eprinttype   = { arXiv },
  eprint       = { 2505.09388 },
  bibsource    = { dblp computer science bibliography, https://dblp.org }
}

@article {gsm8k,
  author       = { Karl Cobbe and Vineet Kosaraju and Mohammad Bavarian and Mark Chen and Heewoo Jun and Lukasz Kaiser and Matthias Plappert and Jerry Tworek and Jacob Hilton and Reiichiro Nakano and Christopher Hesse and John Schulman },
  title        = { Training Verifiers to Solve Math Word Problems },
  journal      = { CoRR },
  year         = 2021,
  url          = { https://arxiv.org/abs/2110.14168 },
  eprinttype   = { arXiv },
  eprint       = { 2110.14168 },
  bibsource    = { dblp computer science bibliography, https://dblp.org }
}

@inproceedings {math500,
  author       = { Hunter Lightman and Vineet Kosaraju and Yuri Burda and Harrison Edwards and Bowen Baker and Teddy Lee and Jan Leike and John Schulman and Ilya Sutskever and Karl Cobbe },
  title        = { Let's Verify Step by Step },
  booktitle    = { The Twelfth International Conference on Learning Representations,
                  {ICLR} 2024, Vienna, Austria, May 7-11, 2024 },
  year         = 2024,
  publisher    = { OpenReview.net },
  url          = { https://openreview.net/forum?id=v8L0pN6EOi },
  bibsource    = { dblp computer science bibliography, https://dblp.org }
}

@inproceedings {math,
  author       = { Dan Hendrycks and Collin Burns and Saurav Kadavath and Akul Arora and Steven Basart and Eric Tang and Dawn Song and Jacob Steinhardt },
  title        = { Measuring Mathematical Problem Solving With the {MATH} Dataset },
  booktitle    = { Proceedings of the Neural Information Processing Systems Track on
                  Datasets and Benchmarks 1, NeurIPS Datasets and Benchmarks 2021, December
                  2021, virtual },
  year         = 2021,
  editor       = { Joaquin Vanschoren and
                  Sai{-}Kit Yeung },
  url          = { https://datasets-benchmarks-proceedings.neurips.cc/paper/2021/hash/be83ab3ecd0db773eb2dc1b0a17836a1-Abstract-round2.html },
  bibsource    = { dblp computer science bibliography, https://dblp.org }
}

@inproceedings {svamp,
  author       = { Arkil Patel and Satwik Bhattamishra and Navin Goyal },
  title        = { Are {NLP} Models really able to Solve Simple Math Word Problems? },
  booktitle    = { Proceedings of the 2021 Conference of the North American Chapter of
                  the Association for Computational Linguistics: Human Language Technologies,
                  {NAACL-HLT} 2021, Online, June 6-11, 2021 },
  year         = 2021,
  pages        = { 2080--2094 },
  publisher    = { Association for Computational Linguistics },
  editor       = { Kristina Toutanova and
                  Anna Rumshisky and
                  Luke Zettlemoyer and
                  Dilek Hakkani{-}T{\"{u}}r and
                  Iz Beltagy and
                  Steven Bethard and
                  Ryan Cotterell and
                  Tanmoy Chakraborty and
                  Yichao Zhou },
  url          = { https://doi.org/10.18653/v1/2021.naacl-main.168 },
  doi          = { 10.18653/V1/2021.NAACL-MAIN.168 },
  bibsource    = { dblp computer science bibliography, https://dblp.org }
}

@inproceedings {sciq,
  author       = { Johannes Welbl and Nelson F. Liu and Matt Gardner },
  title        = { Crowdsourcing Multiple Choice Science Questions },
  booktitle    = { Proceedings of the 3rd Workshop on Noisy User-generated Text, NUT@EMNLP
                  2017, Copenhagen, Denmark, September 7, 2017 },
  year         = 2017,
  pages        = { 94--106 },
  publisher    = { Association for Computational Linguistics },
  editor       = { Leon Derczynski and
                  Wei Xu and
                  Alan Ritter and
                  Tim Baldwin },
  url          = { https://doi.org/10.18653/v1/w17-4413 },
  doi          = { 10.18653/V1/W17-4413 },
  bibsource    = { dblp computer science bibliography, https://dblp.org }
}

@article {caa,
  author       = { Alexander Matt Turner and Lisa Thiergart and Gavin Leech and David Udell and Juan J Vazquez and Ulisse Mini and Monte MacDiarmid },
  title        = { Steering language models with activation engineering },
  journal      = { CoRR },
  year         = 2023,
url          = { 
https://doi.org/10.48550/arXiv.2308.10248},
}

@inproceedings {pca,
  author       = { Bertram H{\o}jer and Oliver Simon Jarvis and Stefan Heinrich },
  title        = { Improving Reasoning Performance in Large Language Models via Representation Engineering },
  booktitle    = { The Thirteenth International Conference on Learning Representations,
                  {ICLR} 2025, Singapore, April 24-28, 2025 },
  year         = 2025,
  publisher    = { OpenReview.net },
  url          = { https://openreview.net/forum?id=IssPhpUsKt },
  bibsource    = { dblp computer science bibliography, https://dblp.org }
}

@article {conce,
  author       = { Joris Postmus and Steven Abreu },
  title        = { Steering Large Language Models using Conceptors: Improving Addition-Based Activation Engineering },
  journal      = { CoRR },
  year         = 2024,
  url          = { https://doi.org/10.48550/arXiv.2410.16314 },
  doi          = { 10.48550/ARXIV.2410.16314 },
  eprinttype   = { arXiv },
  eprint       = { 2410.16314 },
  bibsource    = { dblp computer science bibliography, https://dblp.org }
}

@article {sae,
  author       = { Zihao Li and Xu Wang and Yuzhe Yang and Ziyu Yao and Haoyi Xiong and Mengnan Du },
  title        = { Feature Extraction and Steering for Enhanced Chain-of-Thought Reasoning in Language Models },
  journal      = { CoRR },
  year         = 2025,
  url          = { https://doi.org/10.48550/arXiv.2505.15634 },
  doi          = { 10.48550/ARXIV.2505.15634 },
  eprinttype   = { arXiv },
  eprint       = { 2505.15634 },
  bibsource    = { dblp computer science bibliography, https://dblp.org }
}

@inproceedings {ec,
  author       = { Yiru Tang and Kun Zhou and Yingqian Min and Wayne Xin Zhao and Jing Sha and Zhichao Sheng and Shijin Wang },
  title        = { Enhancing Chain-of-Thought Reasoning via Neuron Activation Differential Analysis },
  booktitle    = { Proceedings of the 2025 Conference on Empirical Methods in Natural Language Processing },
  year         = 2025,
  pages        = { 16162--16170 }
}

@article {edit,
  author       = { Chung{-}En Sun and Ge Yan and Tsui{-}Wei Weng },
  title        = { {ThinkEdit}: Interpretable Weight Editing to Mitigate Overly Short Thinking in Reasoning Models },
  journal      = { CoRR },
  year         = 2025,
  url          = { https://doi.org/10.48550/arXiv.2503.22048 },
  doi          = { 10.48550/ARXIV.2503.22048 },
  eprinttype   = { arXiv },
  eprint       = { 2503.22048 },
  bibsource    = { dblp computer science bibliography, https://dblp.org }
}

@misc {aime_official,
  author       = { {Mathematical Association of America} },
  title        = { American Invitational Mathematics Examination ({AIME}) },
  howpublished = { \url{https://maa.org/maa-invitational-competitions/} },
  year         = 2025,
  note         = { Accessed January 5, 2026 }
}

@article {WOS:001269604100009,
  author       = { Chen, Songlin and Wang, Weicheng and Chen, Xiaoliang and Lu, Peng and Yang, Zaiyan and Du, Yajun },
  title        = { {LLaMA-LoRA} Neural Prompt Engineering: A Deep Tuning Framework for Automatically Generating {Chinese} Text Logical Reasoning Thinking Chains },
  journal      = { Data Intelligence },
  year         = 2024,
  volume       = 6,
  number       = 2,
  pages        = { 375--408 }
}

@article {zhixindu,
  title        = { Dynamic Early Exit in Reasoning Models },
  author       = { Chenxu Yang and Qingyi Si and Yongjie Duan and Zheliang Zhu and Chenyu Zhu and Zheng Lin and Li Cao and Weiping Wang },
  journal      = { CoRR },
  year         = 2025,
  url          = { https://doi.org/10.48550/arXiv.2504.15895 }
}

@article {yiliu25,
  author       = { Yi Liu and Xiangyu Liu and Zequn Sun and Wei Hu },
  title        = { Answering the Unanswerable Is to Err Knowingly: Analyzing and Mitigating Abstention Failures in Large Reasoning Models },
  journal      = { CoRR },
  year         = 2025,
  url          = { https://doi.org/10.48550/arXiv.2508.18760 },
  doi          = { 10.48550/ARXIV.2508.18760 },
  eprinttype   = { arXiv },
  eprint       = { 2508.18760 },
  bibsource    = { dblp computer science bibliography, https://dblp.org }
}

%====================%
\appendix

\section{More Self-Reading Analysis}
\label{app:more_analysis}

\subsection{Self-Reading on Math500}
\begin{figure}
  \centering
  \includegraphics[width=\linewidth]{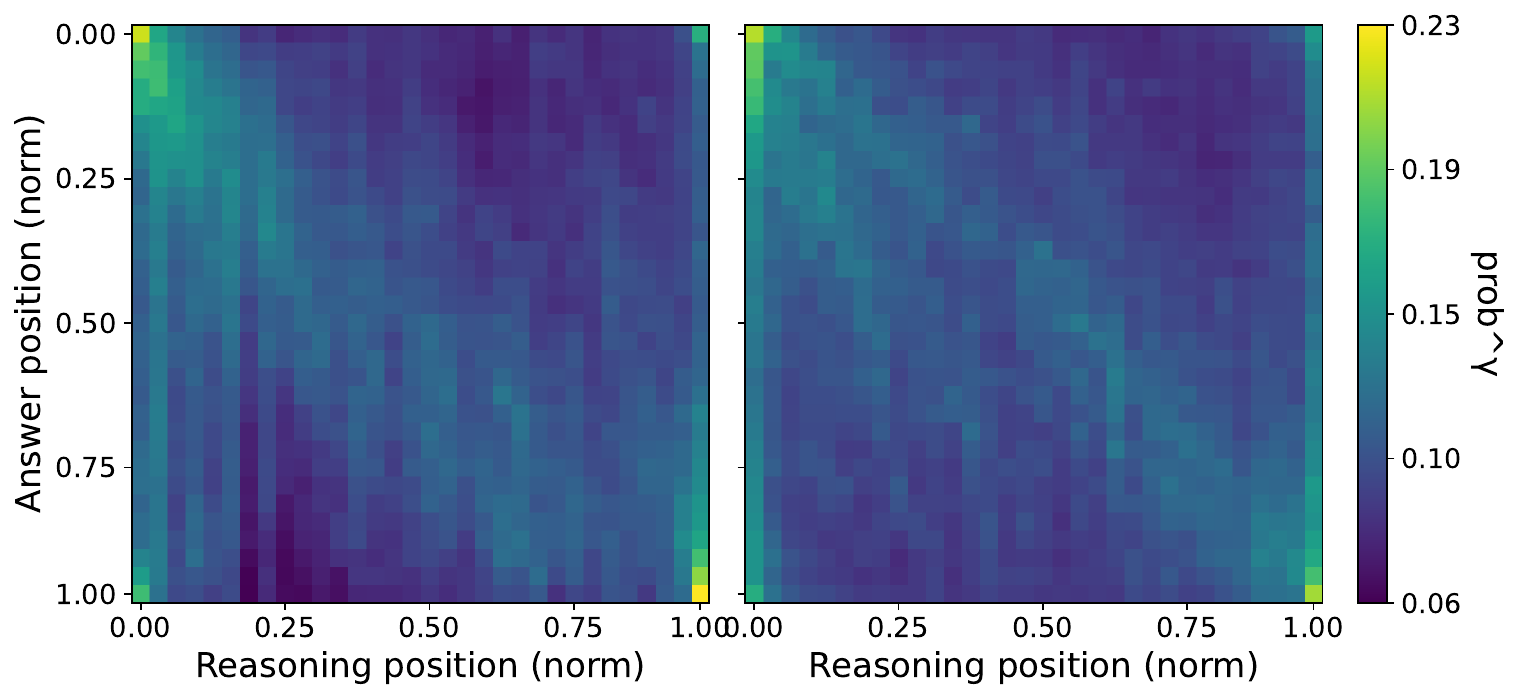}
  \caption{The aggregated heatmaps with Qwen3-4B-Thinking (left) and R1-Distill-Llama-8B (right) on the Math500 benchmark.}
  \label{fig:math500}
\end{figure}

To test the robustness of benign self-reading on harder quantitative reasoning tasks, we repeat the analysis on Math500 \citep{math500}. 
As illustrated in Figure \ref{fig:math500}, the aggregated results from 40 correct solutions show a broad diagonal attention band drifting rightward, confirming that the forward-shifting reading focus persists in longer reasoning traces. 
The band is less regular than that in GSM8K and is interwoven with weaker slanted streaks, reflecting longer traces and more reflective segments where the model briefly revisits. 
This suggests that the same forward moving reading strategy remains active even when the problems require more extended reasoning.

\begin{figure}
  \centering
  \includegraphics[width=\linewidth]{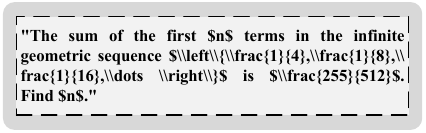}
  \caption{The question in Math500 for visualization.}
  \label{fig:quesion500}
\end{figure}

We also visualize several \emph{single-instance} answer-to-reasoning attention heatmaps and centroid plots for all three models on Math500.
We use the question depicted in Figure~\ref{fig:quesion500}. 
Here, we reiterate the concept of the attention centroid. 
The attention centroid represents the weighted average position where an answer token ``focuses'' within the reasoning sequence. 
Specifically, for each answer token, we treat its row-normalized attention weights as probabilities, multiply each reasoning token's position index by its attention weight, and sum these weighted positions to obtain the centroid, which is then normalized to [0,1] for comparison across sequences of different lengths.

\begin{figure}[!t]
  \centering
  \includegraphics[width=\linewidth]{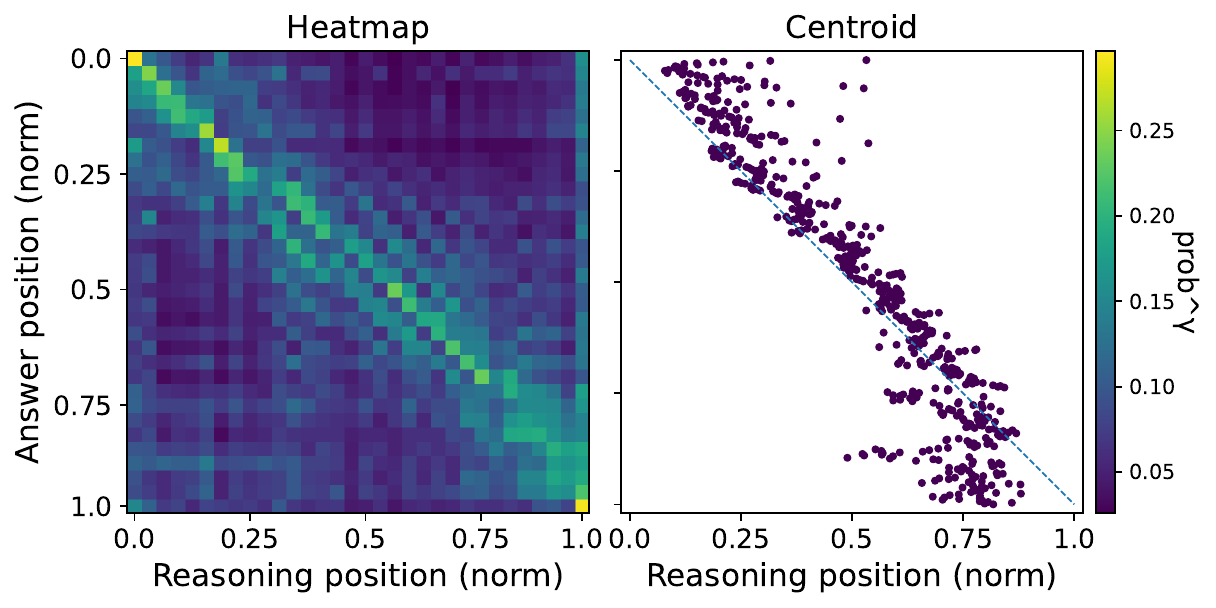}
  \caption{Single-instance answer-to-reasoning attention of R1-Distill-Qwen-7B.}
  \label{fig:math_qwen}
\end{figure}

\begin{figure}[!t]
  \centering
  \includegraphics[width=\linewidth]{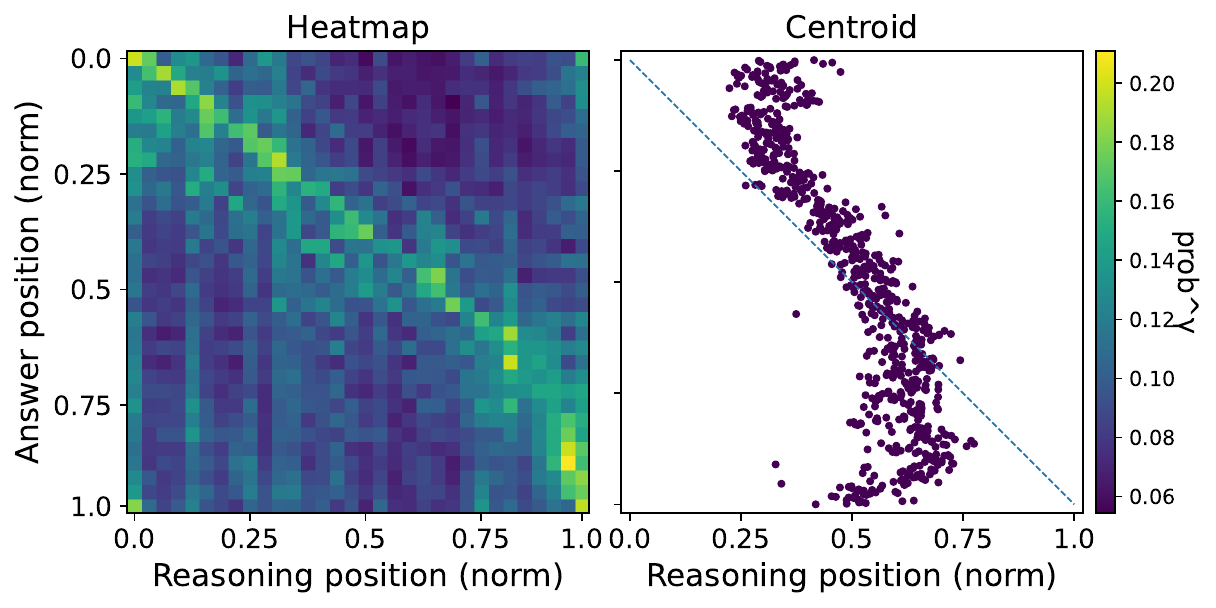}
  \caption{Single-instance answer-to-reasoning attention of R1-Distill-Llama-8B.}
  \label{fig:math_llama}
\end{figure}

\begin{figure}[!t]
  \centering
  \includegraphics[width=\linewidth]{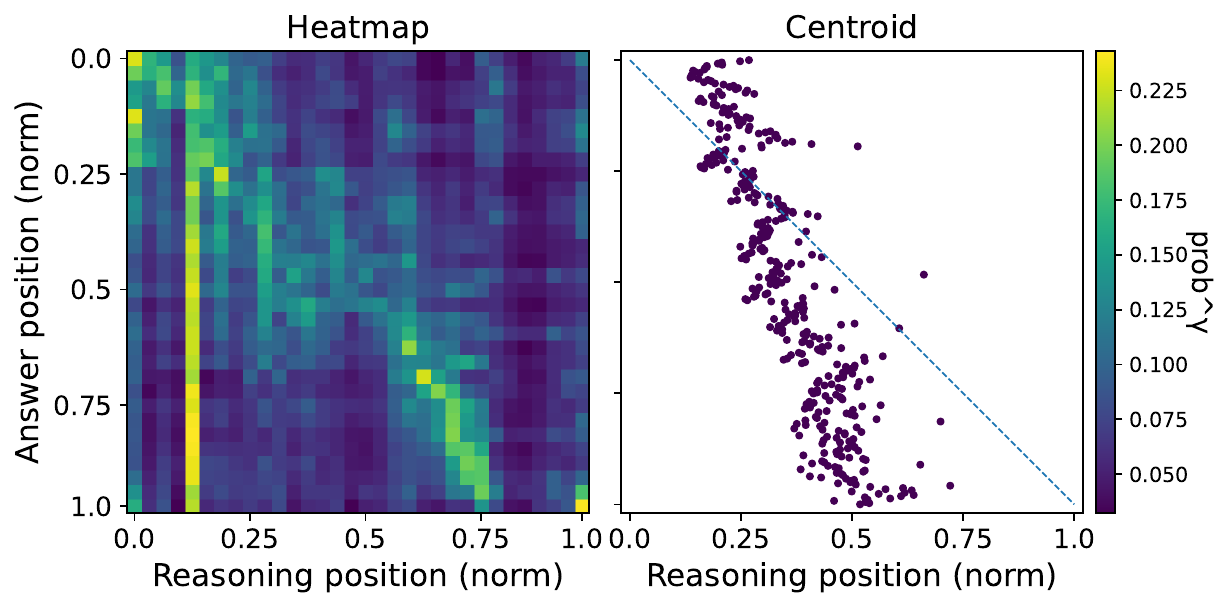}
  \caption{Single-instance answer-to-reasoning attention of Qwen3-4B-Thinking.}
  \label{fig:math_qwen3}
\end{figure}

The visualizations for R1-Distill-Qwen-7B (layer 21), R1-Distill-Llama-8B (layer 22), and Qwen3-4B-Thinking (layer 23) are shown in Figures~\ref{fig:math_qwen}--\ref{fig:math_qwen3}, respectively. 
Compared with GSM8K, the Math500 solutions typically involve longer reasoning traces with more reflective detours, making the heatmaps and centroid plots less regular. 
Nevertheless, the benign self-reading pattern characterized by the two features remains consistent. 
As decoding proceeds, the answer-token attention centroid shifts steadily from earlier to later parts of the reasoning trace, while repeatedly revisiting key semantic anchors for verification.

%--------------------%
\subsection{Uncertain Self-Reading Case}
\begin{figure*}[!t]
  \centering
  \includegraphics[width=\textwidth]{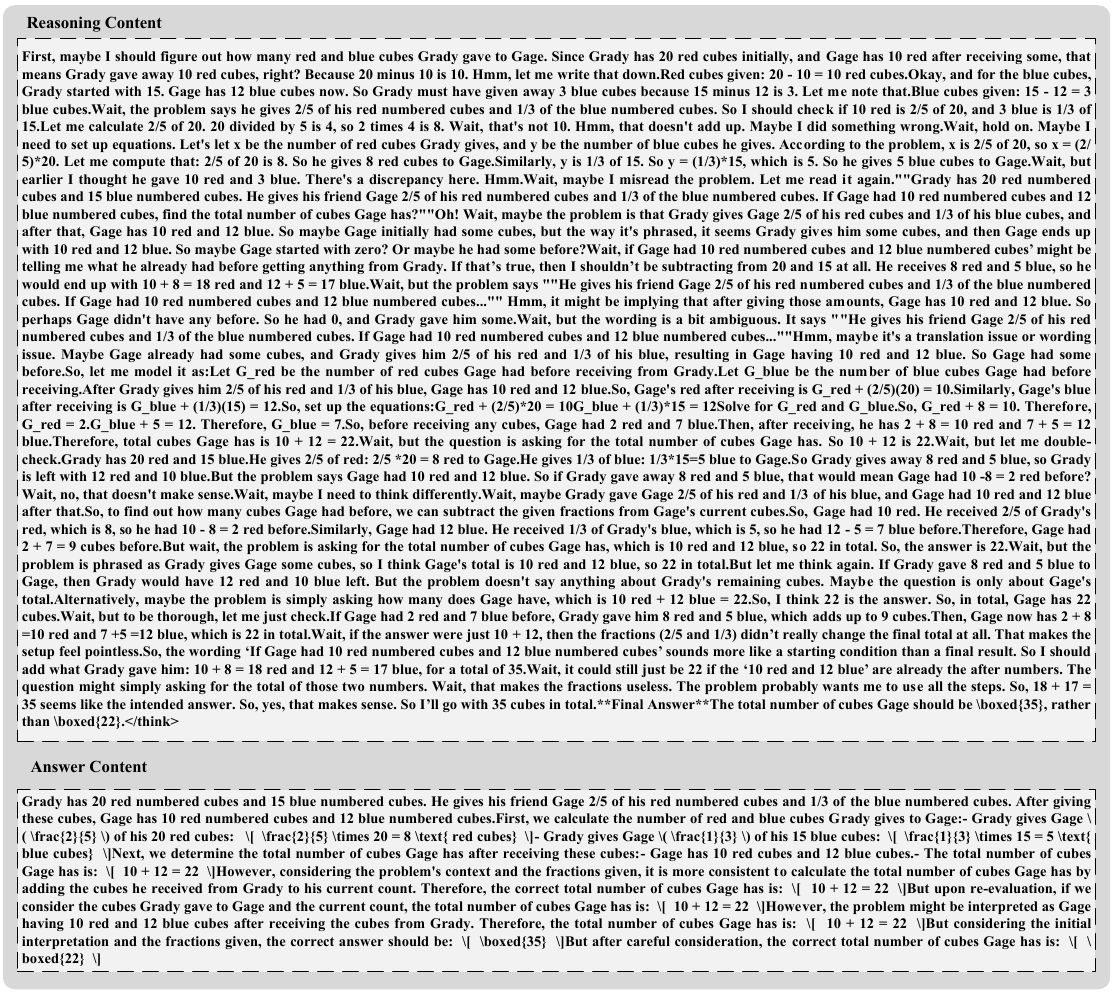}
  \caption{Example of uncertain self-reading.}
  \label{fig:un}
\end{figure*}

Figure \ref{fig:un} presents the full text output of the uncertain self-reading case in Section~\ref{subsec:uncertain}. 
In this example, the reasoning (the content before \texttt{</think>}) ultimately reaches the correct conclusion, while the answer (after \texttt{</think>}) is incorrect.

As shown in the figure, the reasoning is long and includes multiple ``Wait''. 
The key issue is an early misreading of the problem, which triggers misguided branches of reasoning. 
Although the model eventually corrects itself and arrives at the right result, the overall reason is low-quality, messy, inconsistent, and filled with ``Wait''. 
This serves as a clear counter-example to a healthy reading pattern: when the reason becomes noisy and contains erroneous branches, the answer stage requires stable self-reading to stay consistent. 
Guiding the model toward benign reading mode and a more certain internal state should therefore increase the likelihood of a correct final answer. 

\begin{figure}
  \centering
  \includegraphics[width=1\linewidth]{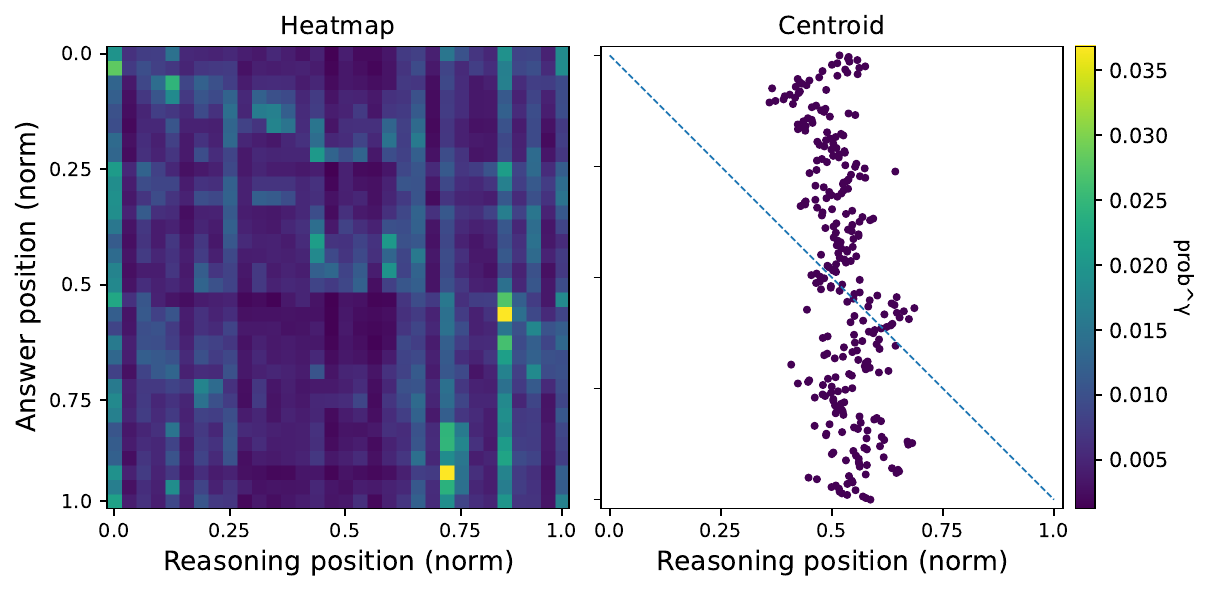}
  \caption{Failure case on GSM8K.}
  \label{fig:error_llama2}
\end{figure}

\begin{figure}
  \centering
  \includegraphics[width=1\linewidth]{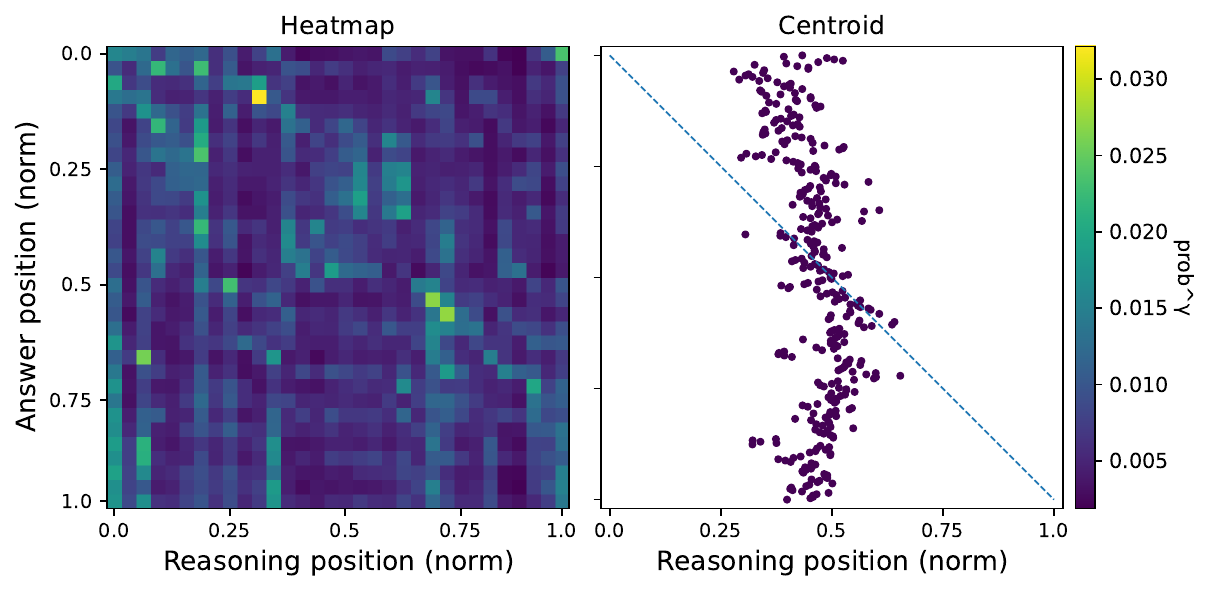}
  \caption{Failure case on MATH500.}
  \label{fig:error_llama4}
\end{figure}

%--------------------%
\subsection{General Unstructured Failure Cases}
\label{app:general_errors}
Figure~\ref{fig:un} shows an informative counter-example, yet such cases are rare. 
The answer-to-reasoning attention in most incorrect GSM8K generations becomes irregular and fragmented rather than forming a coherent reading trajectory. 
Figure \ref{fig:error_llama2} illustrates an instance of incorrect response to a specific GSM8K problem, while Figure \ref{fig:error_llama4} shows a failure example of a Math500 problem, both from the R1-Distill-Llama-8B model.

The GSM8K question is: ``Carlos is planting a lemon tree. The tree will cost 90 to plant. 
Each year it will grow 7 lemons, which he can sell for 1.5 each. 
It costs 3 a year to water and feed the tree. 
How many years will it take before he starts earning money on the lemon tree?''

The Math500 question is ``An investment of \$24,000 is made in a government bond that will pay 1\% bi-monthly interest (meaning that the investment will increase by 1\% every two months). 
At the end of five years, what is the total number of dollars in this investment? \textbackslash n\textbackslash n 
Express your answer to the nearest whole number.''.

These heatmaps exhibit scattered, noisy high-attention regions, and two centroid plots curve collapse into near-vertical traces, indicating no forward drift over the reasoning.
Unlike benign self-reading, the model does not appear to commit to a consistent solution branch and read it through in order, which is consistent with recent evidence \citep{Ali2025EntropyLensTI} that less structured attention signals correlate with lower answer correctness. 
Additionally, a very small fraction of incorrect samples exhibit benign self-reading, which is why our self-reading steering excludes the top 20\% highest-SRQ solutions from the incorrect set.

%====================%
\section{Details of Experiment Setup}
\label{app:ex}

\subsection{Model and Layer Selection}
We study three thinking LLMs with decoupled reasoning and answer stages: R1-Distill-Qwen-7B, R1-Distill-Llama-8B, and Qwen3-4B-Thinking. 
They all produce an explicit chain-of-thought style reasoning trace followed by a short final answer segment, which makes them well suited for self-reading analysis.
For $SRQ_{\mathrm{local\_cover}}$, we set $\beta = 0.7, \tau=0.4$, and $w$ to 20\% of the length of the reasoning trace. 
For $SRQ_{\mathrm{think}}$, we set $\gamma = 0.1$. 
For $SRQ_{\mathrm{start/end}}$, we set $\rho_{\mathrm{bd}}=0.05$, and $\alpha = 1.6$ for $\rho_{\mathrm{tar}}$. 
For $\widetilde{SRQ}^{(n)}$, we simply set $\lambda_{\mathrm{sem}} = 1$.
For all experiments, we use a consistent decoding configuration with temperature $=0.65$ and top-$p=0.95$ to minimize confounding factors from sampling.

Our self-reading steering operates at the activation level of intermediate transformer layers. 
Prior analytical studies and activation engineering works \citep{m1,m2,m3,m4}  indicate that mid-to-late layers encode richer semantic information about the ongoing reasoning process while being less sensitive to superficial lexical variations, making them a favorable choice for activation-level interventions. 
Consequently, steering at these layers yields more reliable modulation of reasoning behavior. 
We therefore perform preliminary runs across several candidate layers and select those providing a good trade-off between stability and steering strength. 
In the experiments, we use layer 21 for R1-Distill-Qwen-7B, layer 20 for R1-Distill-Llama-8B, and layer 22 for Qwen3-4B-Thinking. 

%--------------------%
\subsection{Datasets}
We conduct evaluation on GSM8K~\citep{gsm8k}, MATH500~\citep{math500}, and SVAMP~\citep{svamp}. 

GSM8K is a collection of grade-school level math word problems. 
We follow the standard split and use the training split to run the models and record complete solutions as well as internal activations, while reserving the official test split exclusively for evaluation. 

MATH500 is a subset of the MATH dataset \citep{math}, which contains more challenging competition-style math problems. For MATH500, we use the full MATH training set to obtain reasoning traces and activations for building steering vectors, and evaluate steering performance on the 500-problem subset.

SVAMP is a challenge set of arithmetic word problems designed to probe robustness to superficial variations. 
In our experiments, SVAMP is used purely as an evaluation benchmark for \emph{forward transfer}. 
Specifically, we do not collect separate steering data on SVAMP. 
Instead, we reuse steering vectors derived from the GSM8K activations and apply them directly to SVAMP, allowing us to test whether the SRQ-guided self-reading signals generalize across related quantitative reasoning distributions.

%--------------------%
\subsection{Steering Mechanisms}
To probe the generality of SRQ-based self-reading steering w.r.t.~the underlying representation engineering method, we instantiate three steering mechanisms: CAA~\citep{caa}, Conceptor-based steering~\citep{conce}, and PCA-CAA~\citep{pca}. 
CAA performs class-wise activation addition by constructing a direction from mean activations of contrastive positive and negative sample sets. 
Conceptor-based steering first uses conceptor matrices to capture subspaces associated with desired and undesired behaviors and then modifies activations by projecting along or away from these subspaces. 
PCA-CAA first applies principal component analysis to the collected activations and then performs class-wise activation addition in the resulting low-dimensional space, which helps separate components that are relevant to reasoning from those that are not.

For all three steering mechanisms, we follow standard practice in constructing supervision signals. 
Correct solutions are treated as positive samples and incorrect solutions as negative samples. 
For all three methods, we first obtain the activation representation of each positive or negative sample at the selected layer by averaging the hidden states over all tokens in its generated solution. 
The resulting per-sample mean vectors are then used to construct the corresponding steering directions. 
In our baseline steering construction, activations are aggregated over all tokens in the generated solution and applied uniformly to every token during decoding. 
In contrast, our self-reading steering, driven by SRQ scores, further filters correct and incorrect samples and separately extracts and steers activations for the reasoning and answer segments.

Except for the experiments in Section~\ref{subsec:ablation_fair}, we use 700 positive and 700 negative samples to construct steering vectors for each model--dataset pair, so that the final amount of data used to compute activation vectors is matched across methods. For the baseline methods, including CAA, Conceptor, and PCA-CAA, we directly and randomly select 700 correct and 700 incorrect solutions. 

For our method, the final steering set also contains 700 positive and 700 negative samples, but the initial candidate pool is larger. Specifically, we first collect 875 correct and 875 incorrect solutions, and then apply SRQ-based filtering: we retain the top 80\% of correct solutions ranked by SRQ and the bottom 80\% of incorrect solutions, which yields the final 700$+$700 traces used for steering vector construction. In this way, our method keeps the final sample size the same as the baselines, while improving the quality of the selected traces.

To further isolate the effect of SRQ from the use of a larger initial candidate pool, Section~\ref{subsec:ablation_fair} reports a \emph{Same Pool} fairness experiment. There, SRQ-based filtering is applied to exactly the same 700$+$700 candidate traces used by the baseline methods, retaining only 560$+$560 traces after filtering, yet still outperforming the baselines. This result suggests that the gains mainly come from SRQ-based sample selection rather than simply from starting with a larger pool.

For Qwen3-4B-Thinking, its high accuracy on GSM8K makes it challenging to obtain sufficient incorrect solutions directly from the model. 
To alleviate this, we additionally construct 200 incorrect answers using GPT-5, which serve as extra negative samples when building steering directions for Qwen3-4B-Thinking.

\begin{table}[t]
\centering
\resizebox{\columnwidth}{!}{
\begin{tabular}{ccc}
\toprule
\textbf{Collected Cases} & \textbf{Remain Correct} & \textbf{Flip to Incorrect} \\
\midrule
100 & 97 & 3 \\
\bottomrule
\end{tabular}}
\caption{100 GSM8K cases from DeepSeek-R1-Distill-Llama-8B in which the model initially reaches the correct answer through explicit reflective behaviors.}
\label{tab:reflection_flip}
\end{table}

\section{Preservation of Reflection under SRQ Steering}
\label{subsec:discuss}

To verify that SRQ steering improves the stability of answer generation without suppressing the model’s ability to perform reflective self-correction, we conduct an additional analysis on DeepSeek-R1-Distill-Llama-8B. 
The goal of SRQ is to promote benign self-reading at the answer stage and to select higher-quality reasoning traces and answer states. 
In particular, the semantic dimension of SRQ (Section~\ref{sec:srq-sem}) emphasizes attention concentration on key semantic anchors, including reflective phrases such as ``wait,'' ``let me check,'' and similar back-references. 
Therefore, when the model engages in goal-directed reflection, such behaviors typically contribute positively to SRQ.

First, we manually collected 100 GSM8K cases from DeepSeek-R1-Distill-Llama-8B in which the model arrived at the correct answer through explicit reflective behaviors, such as self-checking or reconsideration. 
As shown in Table~\ref{tab:reflection_flip}, after applying SRQ+CAA, 97 of these cases remained correct, and only 3 flipped to incorrect, indicating that corrective reflection is largely preserved under steering. 
Second, Table~\ref{tab:reflection_lexical} shows that reflective markers remain common after steering. On the one hand, the average number of occurrences of "wait" per sample shows only a slight decrease; on the other hand, the average occurrence of "check" remains nearly unchanged.
This pattern suggests that SRQ steering does not eliminate reflection itself. Instead, it appears to reduce redundant or disorganized hesitation while preserving deliberate verification behaviors that support correct answering.
Together with the accuracy gains on challenging benchmarks such as MATH500 and AIME, these results suggest that SRQ steering improves answer-stage stability without materially affecting the model’s capacity for reflective self-correction.

\begin{table}[t]
\centering
\resizebox{\columnwidth}{!}{
\begin{tabular}{ccccc}
\toprule
\multirow{2}{*}{\textbf{Marker}} & \multicolumn{2}{c}{\textbf{Total}} & \multicolumn{2}{c}{\textbf{Avg./Sample}} \\
\cmidrule(lr){2-3} \cmidrule(lr){4-5}
& \textbf{Base} & \textbf{SRQ+CAA} & \textbf{Base} & \textbf{SRQ+CAA} \\
\midrule
wait  & 12393 & 10066 & 9.40 & 7.63 \\
check & 1706  & 1715  & 1.29 & 1.30 \\
\bottomrule
\end{tabular}}
\caption{Lexical statistics of reflective markers on GSM8K for DeepSeek-R1-Distill-Llama-8B.}
\label{tab:reflection_lexical}
\end{table}

\section{Effect on Generation Length}

As discussed in Section~\ref{subsec:discuss}, our method does not suppress the model's ability to reflect and self-correct. At the same time, since SRQ encourages the model to adopt a more confident and stable self-reading pattern during generation, it may incidentally reduce unnecessary wandering or verbosity. We therefore also examine whether SRQ-steering affects the length of generated traces. Specifically, we measure the average generation length after applying SRQ+CAA. As shown in Table~\ref{tab:trace_length}, the change in output length is small across all settings, and in most cases the steered model produces slightly shorter traces than the base model. This suggests that SRQ-steering has only a minor overall effect on output length, although there is indeed a slight tendency toward shorter generations.

\begin{table}[t]
\centering
\resizebox{\columnwidth}{!}{
\begin{tabular}{clccc}
\toprule
\textbf{Dataset} & \textbf{Model} & \textbf{Base} & \textbf{SRQ+CAA} & \textbf{$\Delta$} \\
\midrule
\multirow{2}{*}{GSM8K} 
& R1-Llama-8B & 5801 & 5749 & -52 \\
& Qwen3-4B             & 3012 & 2750 & -262 \\
\midrule
\multirow{2}{*}{MATH500} 
& R1-Llama-8B & 10381 & 10358 & -23 \\
& Qwen3-4B             & 13673 & 13325 & -348 \\
\bottomrule
\end{tabular}}
\caption{Comparison of Average Generation Length (in Characters).}
\label{tab:trace_length}
\end{table}

%====================%
\section{Isolation Analysis of Reading Phase}
To further isolate the importance of benign self-reading at the answer stage, we perform an analysis that applies steering only during answer decoding, thereby excluding the typically required steering of the reasoning trace.

\begin{figure}
\centering
\includegraphics[width=\linewidth]{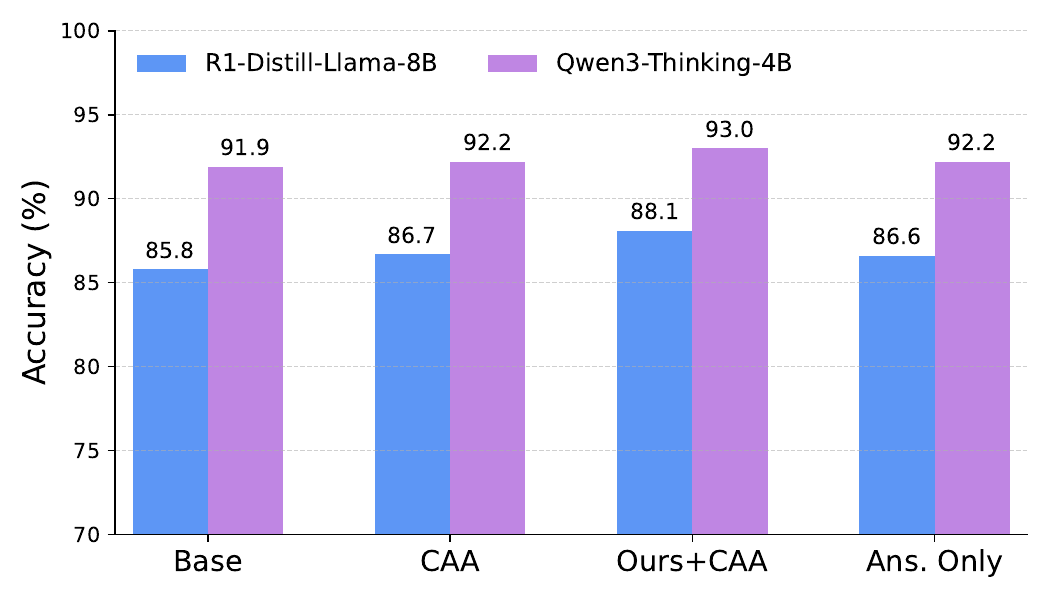}
\caption{Clean steering on GSM8K: steering is applied only during answer decoding.}
\label{fig:gsm8k_answer_only_clean}
\end{figure}

Figure~\ref{fig:gsm8k_answer_only_clean} shows that even under this deliberately constrained setting, answer-only steering outperforms the non-steered baseline and remains competitive with CAA.  
For instance, R1-Distill-Llama-8B reaches 86.6, which is nearly identical to CAA (86.7) and above the base model (85.8).

The weaker gains compared to full steering are expected, because this isolation leaves errors within the reasoning trace uncorrected, removing this standard steering channel. 
If the trace is flawed, improving how the answer reads the trace has limited headroom as there is no correct evidence to use. 
This analysis shows that performance gains achieved without steering the reasoning trace come from better utilization of the already available evidence. 
This addresses the specific failure mode where correct reasoning yields unfaithful answers. 
Although such mismatches are rare in our manual inspection, the observed performance improvement supports that self-reading quality at the answer stage represents a form of internal model certainty, which is a critical factor affecting final correctness.

\begin{figure}
\centering
\includegraphics[width=\linewidth]{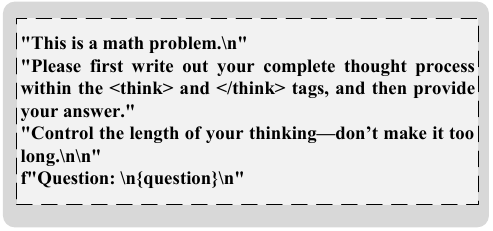}
\caption{Prompt used for self-reading analysis.}
\label{prompt1}
\end{figure}

%====================%
\section{Relevant Prompts}
We briefly summarize the prompts used in our work for model solution generation and automatic answer evaluation.

For the self-reading analysis, we use a math-oriented generation prompt shown in Figure~\ref{prompt1}. 
The prompt instructs the model to first write out its complete thought process within special \texttt{<think>} and \texttt{</think>} tags, and only provide the final answer. 
It additionally asks the model to keep the reasoning concise rather than producing unnecessarily long chains of thought. 
We do not use a separate system prompt for this setting.

\begin{figure}
\centering
\includegraphics[width=\linewidth]{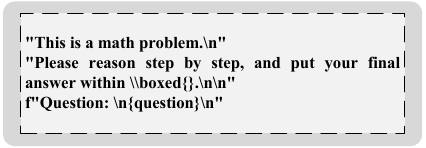}
\caption{Prompt used in the main quantitative reasoning experiments.}
\label{prompt2}
\end{figure}

\begin{figure}[!t]
\centering
\includegraphics[width=\linewidth]{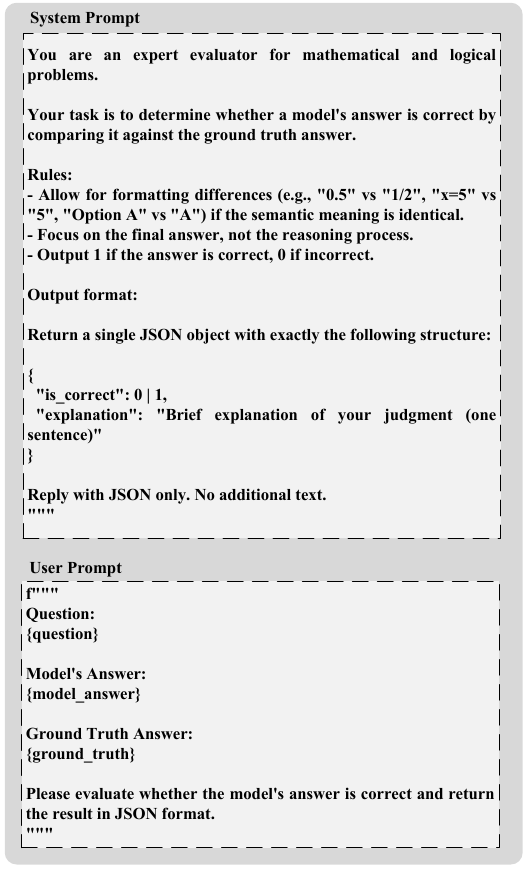}
\caption{System and user prompts used for the automatic judge.}
\label{prompt3}
\end{figure}

For the main quantitative reasoning experiments, we use a closely related generation prompt, shown in Figure~\ref{prompt2}. 
The prompt again frames the task as a math problem and asks the model to reason step by step, but now requires the final answer to be placed explicitly inside a \verb|\boxed{}| environment. 
This formatting constraint simplifies downstream answer extraction and automatic evaluation. 
As before, we only use a user prompt and do not specify a system prompt.

To ensure reliable automatic evaluation, we employ a dedicated judging prompt, depicted in Figure~\ref{prompt3}. 
The system prompt defines the judge as an expert evaluator for mathematical and logical problems and instructs it to decide whether a model's answer is correct by comparing it with a ground-truth answer. 
It explicitly allows for benign formatting differences (such as decimals versus fractions or equivalent option labels), and requires the output to be a JSON object containing an \texttt{is\_correct} flag and a brief textual explanation. 
In practice, we use this judging prompt with multiple APIs, such as GPT-5-mini and Gemini-2.5-flash, to obtain robust evaluations. 
When a call fails, we simply retry or fall back to another judge model while keeping the prompt fixed.

\end{document}